\documentclass[10pt,twocolumn,letterpaper]{article}

\usepackage{cvpr}
\usepackage{times}
\usepackage{epsfig}
\usepackage{graphicx}
\usepackage{amsmath}
\usepackage{amssymb}
\usepackage{enumitem} 
\usepackage{euscript} 
\usepackage{gensymb} 
\usepackage{dsfont} 

\usepackage[pagebackref=true,breaklinks=true,letterpaper=true,colorlinks,bookmarks=false]{hyperref}

\cvprfinalcopy 

\def\cvprPaperID{6545} 

\ifcvprfinal\fi

\long\def\ignorethis#1{}

\usepackage{color}
\definecolor{gray}{rgb}{0.35,0.35,0.35}
\definecolor{red}{rgb}{1,0,0}
\definecolor{dark-green}{rgb}{0,0.4,0}
\definecolor{blue}{rgb}{0,0,1}
\definecolor{orange}{rgb}{1,0.55,0}
\definecolor{white}{rgb}{1,1,1}
\definecolor{black}{rgb}{1,1,1}
\definecolor{dark-brown}{rgb}{0.2,0.1,0}

\newbox\jsavebox

\newcommand{\Lagr}{\mathcal{L}}
\newcommand{\SPNet}{SampleNet }
\newcommand{\SPNetwo}{SampleNet}

\DeclareMathOperator*{\argmax}{\arg\!\max}
\DeclareMathOperator*{\argmin}{\arg\!\min}

\begin{document}

\title{\SPNetwo: Differentiable Point Cloud Sampling}

\author{Itai Lang\\
Tel Aviv University\\
{\tt\small itailang@mail.tau.ac.il}
\and
Asaf Manor\\
Tel Aviv University\\
{\tt\small asafmanor@mail.tau.ac.il}
\and
Shai Avidan\\
Tel Aviv University\\
{\tt\small avidan@eng.tau.ac.il}
}
\maketitle
\thispagestyle{empty} 

\begin{abstract}
There is a growing number of tasks that work directly on point clouds. As the size of the point cloud grows, so do the computational demands of these tasks. A possible solution is to sample the point cloud first. Classic sampling approaches, such as farthest point sampling (FPS), do not consider the downstream task. A recent work showed that learning a task-specific sampling can improve results significantly. However, the proposed technique did not deal with the non-differentiability of the sampling operation and offered a workaround instead.

We introduce a novel differentiable relaxation for point cloud sampling that approximates sampled points as a mixture of points in the primary input cloud. Our approximation scheme leads to consistently good results on classification and geometry reconstruction applications. We also show that the proposed sampling method can be used as a front to a point cloud registration network. This is a challenging task since sampling must be consistent across two different point clouds for a shared downstream task. In all cases, our approach outperforms existing non-learned and learned sampling alternatives. Our code is publicly available\footnote{\url{https://github.com/itailang/SampleNet}}. 


\end{abstract}

\section{Introduction} \label{sec:introduction}

\begin{figure}[t!]
\begin{center}
\includegraphics[width=\columnwidth]{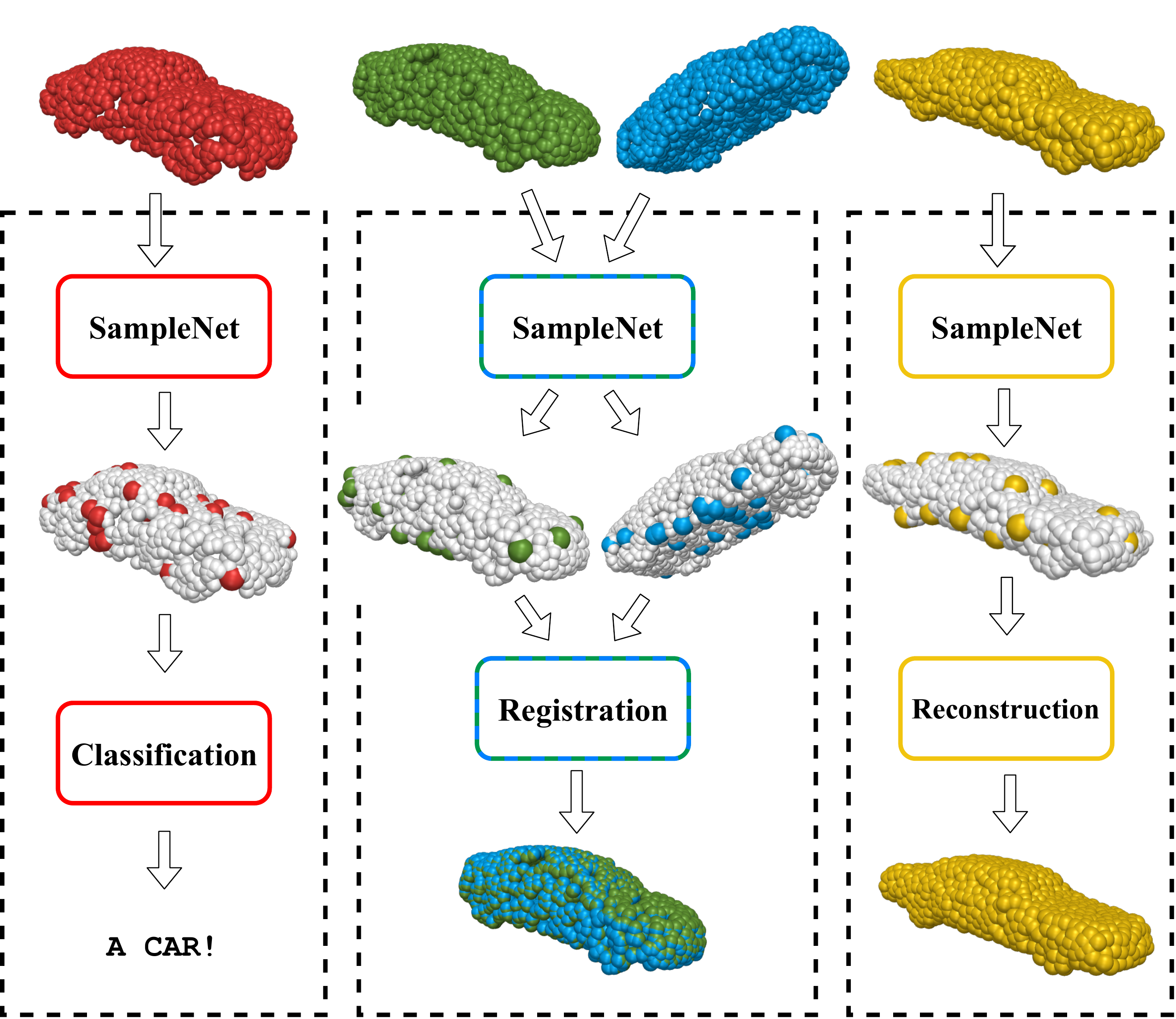}
\caption{{\bfseries Applications of \SPNetwo.} Our method learns to sample a point cloud for a subsequent task. It employs a differentiable relaxation of the selection of points from the input point cloud. \SPNet lets various tasks, such as classification, registration, and reconstruction, to operate on a small fraction of the input points with minimal degradation in performance.}
\label{fig:teaser}
\end{center}
\end{figure}

The popularity of 3D sensing devices increased in recent years. These devices usually capture data in the form of a point cloud - a set of points representing the visual scene. A variety of applications, such as classification, registration and shape reconstruction, consume the raw point cloud data. These applications can digest large point clouds, though it is desirable to reduce the size of the point cloud (Figure~\ref{fig:teaser}) to improve computational efficiency and reduce communication costs. 

This is often done by sampling the data before running the downstream task \cite{gelfand2003geometrically, himmelsbach2009real, klokov2017escape}. Since sampling preserves the data structure (i.e., both input and output are point clouds), it can be used natively in a process pipeline. Also, sampling preserves data fidelity and retains the data in an interpretable representation.


An emerging question is how to select the data points. A widely used method is farthest point sampling (FPS)~\cite{qi2017pointnetplusplus, yu2018pu-net, li2018pointcnn, qi2019deep}. FPS starts from a point in the set, and iteratively selects the farthest point from the points already selected~\cite{eldar1997FPS, moenning2003fast}. It aims to achieve a maximal coverage of the input.

FPS is task agnostic. It minimizes a geometric error and does not take into account the subsequent processing of the sampled point cloud. A recent work by Dovrat \etal~\cite{dovrat2019learning} presented a task-specific sampling method. Their key idea was to simplify and then sample the point cloud. In the first step, they used a neural network to produce a small set of simplified points in the ambient space, optimized for the task. This set is not guaranteed to be a subset of the input. Thus, in a post-processing step, they matched each simplified point to its nearest neighbor in the input point cloud, which yielded a subset of the input.


This learned sampling approach improved application performance with sampled point clouds, in comparison to non-learned methods, such as FPS and random sampling. However, the matching step is a non-differentiable operation and can not propagate gradients through a neural network. This substantially compromises the performance with sampled points in comparison to the simplified set, since matching was not introduced at the training phase.

\begin{figure}[tb!]
\includegraphics[width=1\columnwidth]{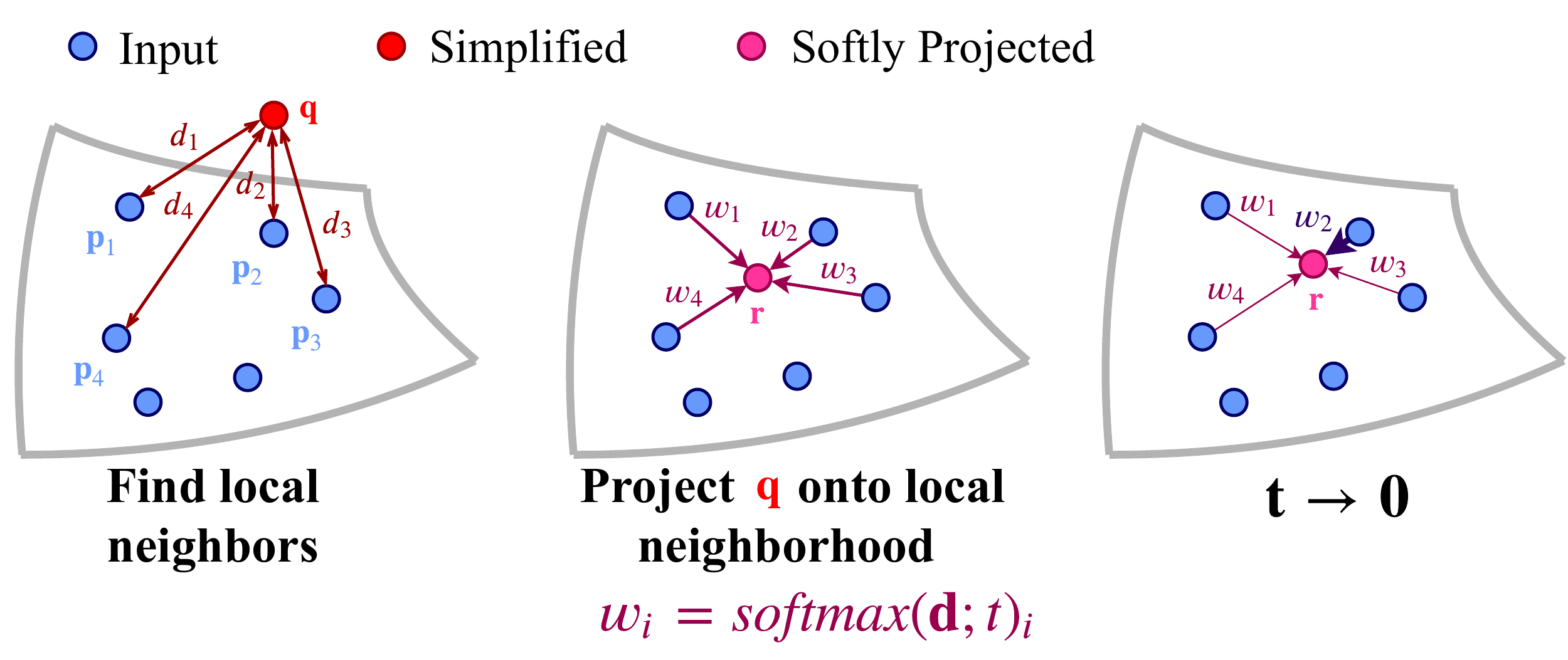}
\caption{{\bfseries Illustration of the sampling approximation.} We propose a learned sampling approach for point clouds that employs a differentiable relaxation to nearest neighbor selection. A query point $\mathbf{q}$ (in Red) is projected onto its local neighborhood from the input point cloud (in Blue). A weighted average of the neighbors form a softly projected point $\mathbf{r}$ (in Magenta). During training the weights are optimized to approximated nearest neighbor sampling ($\mathbf{p}_2$ in this example), which occurs at inference time.}
\label{fig:flow}
\end{figure}

We extend the work of Dovrat \etal~\cite{dovrat2019learning} by introducing a differentiable relaxation to the matching step, i.e., nearest neighbor selection, during training (Figure~\ref{fig:flow}). This operation, which we call {\em soft projection}, replaces each point in the simplified set with a weighted average of its nearest neighbors from the input. During training, the weights are optimized to approximate the nearest neighbor selection, which is done at inference time.

The soft projection operation makes a change in representation. Instead of absolute coordinates in the free space, the projected points are represented in weight coordinates of their local neighborhood in the initial point cloud. The operation is governed by a temperature parameter, which is minimized during the training process to create an annealing schedule~\cite{laarhoven1987simulated}. The representation change renders the optimization goal as multiple localized classification problems, where each simplified point should be assigned to an optimal input point for the subsequent task.


Our method, termed \SPNetwo, is applied to a variety of tasks, as demonstrated in Figure~\ref{fig:teaser}. Extensive experiments show that we outperform the work of Dovrat \etal consistently. Additionally, we examine a new application - registration with sampled point clouds and show the advantage of our method for this application as well. Registration introduces a new challenge: the sampling algorithm is required to sample consistent points across two different point clouds for a common downstream task. To summarize, our key contributions are threefold:
\begin{itemize}[noitemsep, nolistsep]
    \item A novel differentiable approximation of point cloud sampling;
    \item Improved performance with sampled point clouds for classification and reconstruction tasks, in comparison to non-learned and learned sampling alternatives;
    \item Employment of our method for point cloud registration.
\end{itemize}

\section{Related Work} \label{sec:related_work}

\noindent {\bfseries Deep learning on point clouds} \quad Early research on deep learning for 3D point sets focused on regular representations of the data, in the form of 2D multi-views~\cite{qi2016volumetric, su2017multi} or 3D voxels~\cite{wu2015modelnet, qi2016volumetric}. These representations enabled the natural extension of successful neural processing paradigms from the 2D image domain to 3D data. However, point clouds are irregular and sparse. Regular representations come with the cost of high computational load and quantization errors.

PointNet~\cite{qi2017pointnet} pioneered the direct processing of raw point clouds. It includes per point multi-layer perceptrons (MLPs) that lift each point from the coordinate space to a high dimensional feature space. A global pooling operation aggregates the information to a representative feature vector, which is mapped by fully connected (FC) layers to the object class of the input point cloud.

The variety of deep learning applications for point clouds expanded substantially in the last few years. Today, applications include point cloud classification~\cite{qi2017pointnetplusplus, li2018pointcnn, thomas2019kpconv, wu2019pointconv}, part segmentation~\cite{li2018so-net, su2018splatnet, liu2019relation, wang2019dynamic}, instance segmentation~\cite{wang2018sgpn, yi2019gspn, wang2019associatively}, semantic segmentation~\cite{landrieu2018large, pham2019jsis3d, wang2019graph}, and object detection in point clouds~\cite{qi2019deep, shi2019pointrcnn}. Additional applications include point cloud autoencoders~\cite{achlioptas2018learning, yang2018folding, han2019multi, zhao20193dpoint}, point set completion~\cite{yuan2018pcn, chen2019unpaired, sarmad2019rl-gan-net} and registration~\cite{aoki2019pointnetlk, lu2019deepicp, sarode2019pcrnet}, adversarial point cloud generation~\cite{li2018pointcloudgan, yang2019pointflow}, and adversarial attacks~\cite{liu2019extending, xiang2019generating}. Several recent works studied the topic of point cloud consolidation~\cite{yu2018pu-net, yu2018ec-net, li2019pu-gan, yifan2019patch}. Nevertheless, little attention was given to sampling strategies for point sets.

\medskip 
\noindent {\bfseries Nearest neighbor selection} \quad Nearest neighbor (NN) methods have been widely used in the literature for information fusion~\cite{goldberger2005neighbourhood, qi2017pointnetplusplus, plotz2018neural, wang2019dynamic}. A notable drawback of using nearest neighbors, in the context of neural networks, is that the selection rule is non-differentiable. Goldberger \etal~\cite{goldberger2005neighbourhood} suggested a stochastic relaxation of the nearest neighbor rule. They defined a categorical distribution over the set of candidate neighbors, where the 1-NN rule is a limit case of the distribution.

Later on, Pl\"{o}tz and Roth~\cite{plotz2018neural} generalized the work of Goldberger \etal, by presenting a deterministic relaxation of the $k$ nearest neighbor (KNN) selection rule. They proposed a neural network layer, dubbed neural nearest neighbors block, that employs their KNN relaxation. In this layer, a weighted average of neighbors in the features space is used for information propagation. The neighbor weights are scaled with a temperature coefficient that controls the uniformity of the weight distribution. In our work, we employ the relaxed nearest neighbor selection as a way to approximate point cloud sampling. While the temperature coefficient is unconstrained in the work of Pl\"{o}tz and Roth, we promote a small temperature value during training, to approximate the nearest neighbor selection.



\medskip
\noindent {\bfseries Sampling methods for points clouds in neural networks} \quad Farthest point sampling (FPS) has been widely used as a pooling operation in point cloud neural processing systems \cite{qi2017pointnetplusplus, qi2019deep, yin2019logan}. However, FPS does not take into account the further processing of the sampled points and may result in sub-optimal performance. Recently, alternative sub-sampling methods have been proposed~\cite{li2019unsupervised, nezhadarya2019adaptive, yang2019modeling}. Nezhadarya \etal~\cite{nezhadarya2019adaptive} introduced a critical points layer, which passes on points with the most active features to the next network layer. Yang \etal~\cite{yang2019modeling} used Gumbel subset sampling during the training of a classification network instead of FPS, to improve its accuracy. The settings of our problem are different though. Given an application, we sample the input point cloud and apply the task on the sampled data.

Dovrat \etal~\cite{dovrat2019learning} proposed a learned task-oriented simplification of point clouds, which led to a performance gap between train and inference phases. We mitigate this problem by approximating the sampling operation during training, via a differentiable nearest neighbor approximation.

\section{Method}  \label{sec:method}
\begin{figure}[tb!]
\includegraphics[width=\columnwidth]{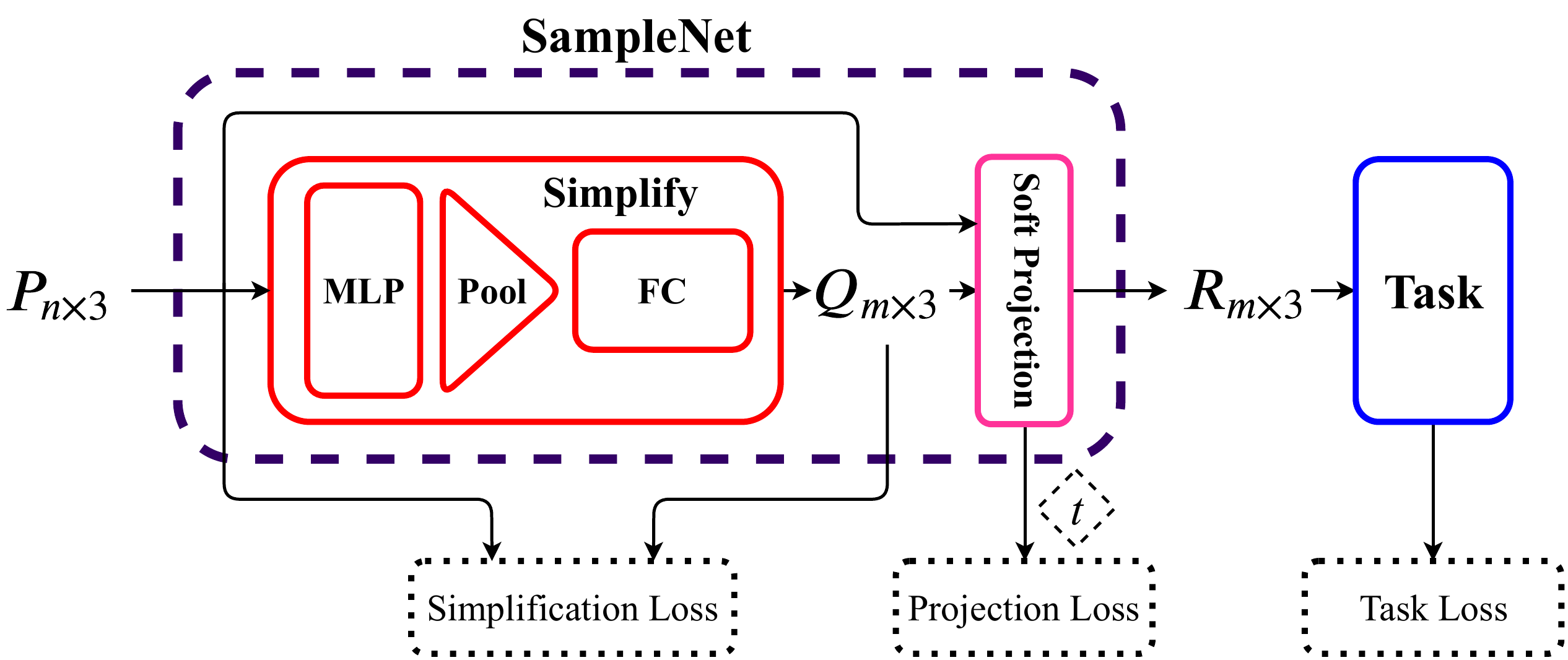}
\caption{{\bfseries Training of the proposed sampling method.} The task network trained on complete input point clouds $P$ and kept fixed during the training of our sampling network \SPNetwo. $P$ is simplified with a neural network to a smaller set $Q$. Then, $Q$ is softly projected onto $P$ to obtain $R$, and $R$ is fed to the task network. Subject to the denoted losses, \SPNet is trained to sample points from $P$ that are optimal for the task at hand.}
\label{fig:system_overview}
\end{figure}

An overview of our sampling method, \SPNetwo, is depicted in Figure~\ref{fig:system_overview}. First, a task network is pre-trained on complete point clouds of $n$ points and frozen. Then, \SPNet takes a complete input $P$ and simplifies it via a neural network to a smaller set $Q$ of $m$ points~\cite{dovrat2019learning}. $Q$ is soft projected onto $P$ by a differentiable relaxation of nearest neighbor selection. Finally, the output of \SPNetwo, $R$, is fed to the task.

\SPNet is trained with three loss terms:
\begin{equation} \label{eq:loss_total_sample}
\begin{split}
\Lagr_{total}^{samp} = \Lagr_{task}(R) &+ \alpha \Lagr_{simplify}(Q,P) \\
                                       &+ \lambda \Lagr_{project}.
\end{split}
\end{equation}

\noindent The first term, $\Lagr_{task}(R)$, optimizes the approximated sampled set $R$ to the task. It is meant to preserve the task performance with sampled point clouds. $\Lagr_{simplify}(Q,P)$ encourages the simplified set to be close to the input. That is, each point in $Q$ should have a close point in $P$ and vice-versa. The last term, $\Lagr_{project}$ is used to approximate the sampling of points from the input point cloud by the soft projection operation.

\begin{figure}[tb!]
\includegraphics[width=\columnwidth]{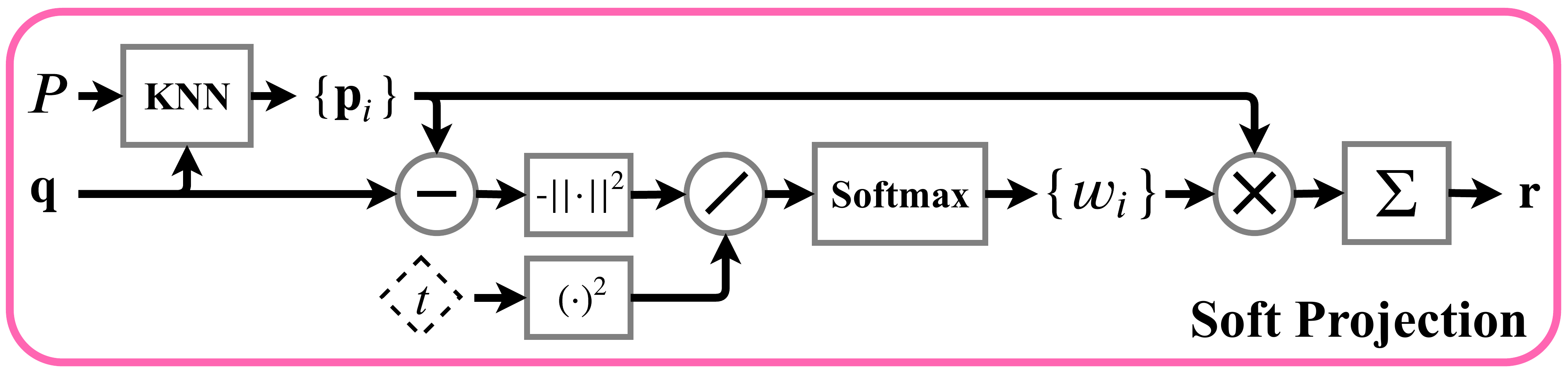}
\caption{{\bfseries The soft projection operation.} The operation gets as input the point cloud $P$ and the simplified point cloud $Q$. Each point $\mathbf{q} \in Q$ is projected onto its $k$ nearest neighbors in $P$, denoted as $\{\mathbf{p}_i\}$. The neighbors $\{\mathbf{p}_i\}$ are weighted by $\{w_i\}$, according to their distance from $\mathbf{q}$ and a temperature coefficient $t$, to obtain a point $\mathbf{r}$ in the soft projected point set $R$.}
\label{fig:project}
\end{figure}

Our method builds on and extends the sampling approach proposed by Dovrat \etal~\cite{dovrat2019learning}. For clarity, we briefly review their method in section~\ref{sec:Simplify}. Then, we describe our extension in section~\ref{sec:Project}.

\subsection{Simplify} \label{sec:Simplify}
Given a point cloud of $n$ 3D coordinates $P \in \mathbb{R}^{n \times 3}$, the goal is to find a subset of $m$ points $R^* \in \mathbb{R}^{m \times 3}$, such that the sampled point cloud $R^*$ is optimized to a task $T$. Denoting the objective function of $T$ as $\EuScript{F}$, $R^*$ is given by:
\begin{equation} \label{eq:problem_statement}
R^* = \argmin_R \EuScript{F}(T(R)), \quad R \subseteq P, \quad |R|=m \leq n.
\end{equation}

This optimization problem poses a challenge due to the non-differentiability of the sampling operation. Dovrat \etal~\cite{dovrat2019learning} suggested a simplification network that produces $Q$ from $P$, where $Q$ is optimal for the task and its points are close to those of $P$. In order to encourage the second property, a simplification loss is utilized. Denoting average nearest neighbor loss as:
\begin{equation} \label{eq:loss_a}
\Lagr_a(X,Y) = \frac{1}{|X|}\sum_{\mathbf{x} \in X}{\min_{\mathbf{y} \in Y}||\mathbf{x}-\mathbf{y}||_2^2},
\end{equation}

\noindent and maximal nearest neighbor loss as:
\begin{equation} \label{eq:loss_m}
\Lagr_m(X,Y) = \max_{\mathbf{x} \in X}{\min_{\mathbf{y} \in Y}||\mathbf{x}-\mathbf{y}||_2^2},
\end{equation}

\noindent the simplification loss is given by:
\begin{equation} \label{eq:loss_simplify}
\begin{split}
\Lagr_{simplify}(Q,P) = \Lagr_a(Q,P) + \beta \Lagr_m(Q,P)& \\
         + (\gamma + \delta|Q|) \Lagr_a(P,Q)&.
\end{split}
\end{equation}

In order to optimize the point set $Q$ to the task, the task loss is added to the optimization objective. The total loss of the simplification network is:
\begin{equation} \label{eq:loss_s}
\Lagr_{s}(Q,P)= \Lagr_{task}(Q) + \alpha \Lagr_{simplify}(Q,P).
\end{equation}

The simplification network described above is trained for a specific sample size $m$. Dovrat \etal~\cite{dovrat2019learning} also proposed a progressive sampling network. This network orders the simplified points according to their importance for the task and can output any sample size. It outputs $n$ points and trained with simplification loss on nested subsets of its output:
\begin{equation} \label{eq:loss_progressive}
\Lagr_{prog}(Q,P)= \sum_{c \in C_s}{\Lagr_{s}(Q_c,P)},
\end{equation}

\noindent where $C_s$ are control sizes.

\subsection{Project} \label{sec:Project}

Instead of optimizing the simplified point cloud for the task, we add the {\em soft projection} operation. The operation is depicted in Figure~\ref{fig:project}. Each point $\mathbf{q} \in Q$ is softly projected onto its neighborhood, defined by its $k$ nearest neighbors in the complete point cloud $P$, to obtain a projected point $\mathbf{r} \in R$. The point $\mathbf{r}$ is a weighted average of original points form $P$:
\begin{equation} \label{eq:soft_proj}
\mathbf{r} = \sum_{i \in \EuScript{N}_P(\mathbf{q})}{w_i \mathbf{p}_i},
\end{equation}

\noindent where $\EuScript{N}_P(\mathbf{q})$ contains the indices of the $k$ nearest neighbors of $\mathbf{q}$ in $P$. The weights $\{w_i\}$ are determined according to the distance between $\mathbf{q}$ and its neighbors, scaled by a learnable temperature coefficient $t$:
\begin{equation} \label{eq:w_i}
w_i = \frac{e^{-d_i^2/t^2}}{\sum_{j \in \EuScript{N}_P(\mathbf{q})}e^{-d_j^2/t^2}},
\end{equation}

\noindent The distance is given by $d_i = ||\mathbf{q}-\mathbf{p}_i||_2$.

The neighborhood size $k = |\EuScript{N}_P(\mathbf{q})|$ plays a role in the choice of sampled points. Through the distance terms, the network can adapt a simplified point's location such that it will approach a different input point in its local region. While a small neighborhood size demotes exploration, choosing an excessive size may result in loss of local context.

The weights $\{w_i\}$ can be viewed as a probability distribution function over the points $\{\mathbf{p}_i\}$, where $\mathbf{r}$ is the expectation value. The temperature coefficient controls the shape of this distribution. In the limit of $t \rightarrow 0$, the distribution converges to a Kronecker delta function, located at the nearest neighbor point.

Given these observations, we would like the point $\mathbf{r}$ to approximate nearest neighbor sampling from the local neighborhood in $P$. To achieve this we add a projection loss, given by:
\begin{equation} \label{eq:loss_project}
\Lagr_{project} = t^2.
\end{equation}

\noindent This loss promotes a small temperature value.

In our sampling approach, the task network is fed with the projected point set $R$ rather than simplified set $Q$. Since each point in $R$ estimates the selection of a point from $P$, our network is trained to {\em sample} the input point cloud rather than simplify it.

Our sampling method can be easily extended to the progressive sampling settings (Equation~\ref{eq:loss_progressive}). In this case, the loss function takes the form:
\begin{equation} \label{eq:loss_total_progressive}
\begin{split}
\Lagr_{total}^{prog} = \sum_{c \in C_s}(\Lagr_{task}(R_c)
                     &+ \alpha \Lagr_{simplify}(Q_c,P)) \\
                     &+ \lambda \Lagr_{project},
\end{split}
\end{equation}

\noindent where $R_c$ is the point set obtained by applying the soft projection operation on $Q_c$ (Equation~\ref{eq:soft_proj}).

At inference time we replace the soft projection with sampling, to obtain a sampled point cloud $R^*$. Like in a classification problem, for each point $\mathbf{r}^* \in R^*$, we select the point $\mathbf{p}_i$ with the highest projection weight:
\begin{equation} \label{eq:hard_proj}
\mathbf{r}^* = \mathbf{p}_{i^*}, \quad i^* = \argmax_{i \in \EuScript{N}_P(\mathbf{q})} w_i.
\end{equation}


Similar to Dovrat \etal~\cite{dovrat2019learning}, if more than one point $\mathbf{r}^*$ corresponds the same point $\mathbf{p}_{i^*}$, we take the unique set of sampled points, complete it using FPS up to $m$ points and evaluate the task performance.

\medskip
\noindent {\bfseries Soft projection as an idempotent operation} \quad
Strictly speaking, the soft projection operation (Equation~\ref{eq:soft_proj}) is not idempotent \cite{valenza2012linear} and thus does not constitute a mathematical projection. However, when the temperature coefficient in Equation~\ref{eq:w_i} goes to zero, the idempotent sampling operation is obtained (Equation~\ref{eq:hard_proj}). Furthermore, the nearest neighbor selection can be viewed as a variation of projection under the Bregman divergence~\cite{chen2008metrics}. The derivation is given in the supplementary.

\section{Results}\label{sec:resutls}

In this section, we present the results of our sampling approach for various applications: point cloud classification, registration, and reconstruction. The performance with point clouds sampled by our method is contrasted with the commonly used FPS and the learned sampling method, S-NET, proposed by Dovrat \etal~\cite{dovrat2019learning}.

Classification and registration are benchmarked on ModelNet40~\cite{wu2015modelnet}. We use point clouds of 1024 points that were uniformly sampled from the dataset models. The official train-test split~\cite{qi2017pointnet} is used for training and evaluation.

The reconstruction task is evaluated with point sets of 2048 points, sampled from ShapeNet Core55 database~\cite{chang2015shapenet}. We use four shape classes with the largest number of examples: Table, Car, Chair, and Airplane. Each class is split to 85\%/5\%/10\% for train/validation/test sets.

Our network \SPNet is based on PointNet architecture. It operates directly on point clouds and is invariant to permutations of the points. \SPNet applies MLPs to the input points, followed by a global max pooling. Then, a simplified point cloud is computed from the pooled feature vector and projected onto the input point cloud. The complete experimental settings are detailed in the supplemental.

\subsection{Classification} \label{sec:classification}
Following the experiment of Dovrat \etal~\cite{dovrat2019learning}, we use PointNet~\cite{qi2017pointnet} as the task network for classification. PointNet is trained on point clouds of 1024 points. Then, instance classification accuracy is evaluated on sampled point clouds from the official test split. The sampling ratio is defined as $1024/m$, where $m$ is the number of sampled points.

\begin{figure}[htb!]
\includegraphics[width=\columnwidth]{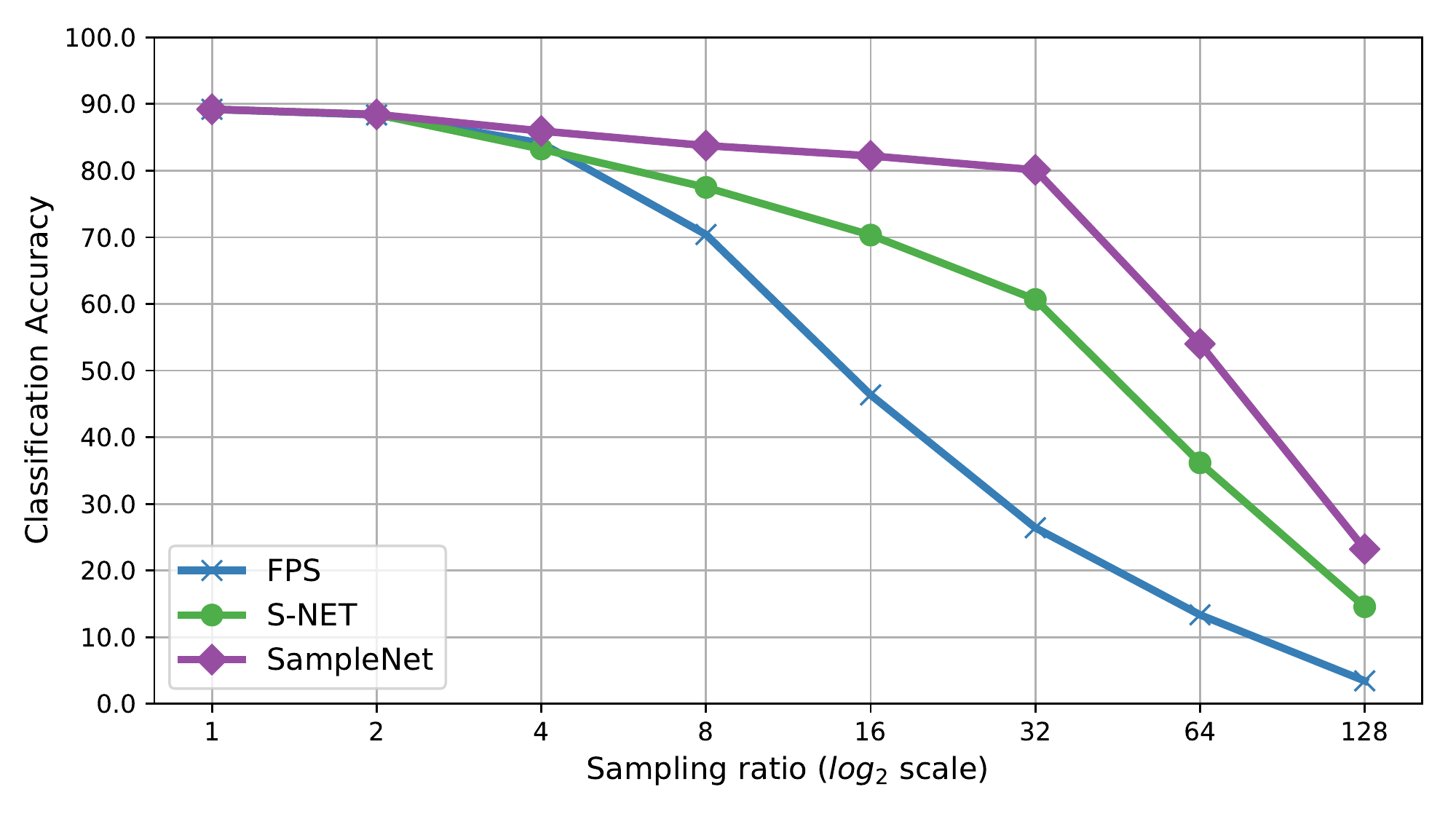}
\caption{{\bfseries Classification accuracy with \SPNetwo.} PointNet is used as the task network and was pre-trained on complete point clouds with 1024 points. The instance classification accuracy is evaluated on sampled point clouds from the test split of ModelNet40. Our sampling method \SPNet outperforms the other sampling alternatives with a large gap.}
\label{fig:SPNet_single_classification}
\end{figure}

\medskip
\noindent {\bfseries \SPNetwo} \quad Figure~\ref{fig:SPNet_single_classification} compares the classification performance for several sampling methods. FPS is agnostic to the task, thus leads to substantial accuracy degradation as the sampling ratio increases. S-NET improves over FPS. However, S-NET is trained to simplify the point cloud, while at inference time, sampled points are used. Our \SPNet is trained directly to sample the point cloud, thus, outperforms the competing approaches by a large margin.

For example, at sampling ratio 32 (approximately 3\% of the original points), it achieves 80.1\% accuracy, which is 20\% improvement over S-NET's result and only 9\% below the accuracy when using the complete input point set. \SPNet also achieves performance gains with respect to FPS and S-NET in progressive sampling settings (Equation~\ref{eq:loss_progressive}). Results are given in the supplementary material.

\begin{figure}[tb!]
\includegraphics[width=\columnwidth]{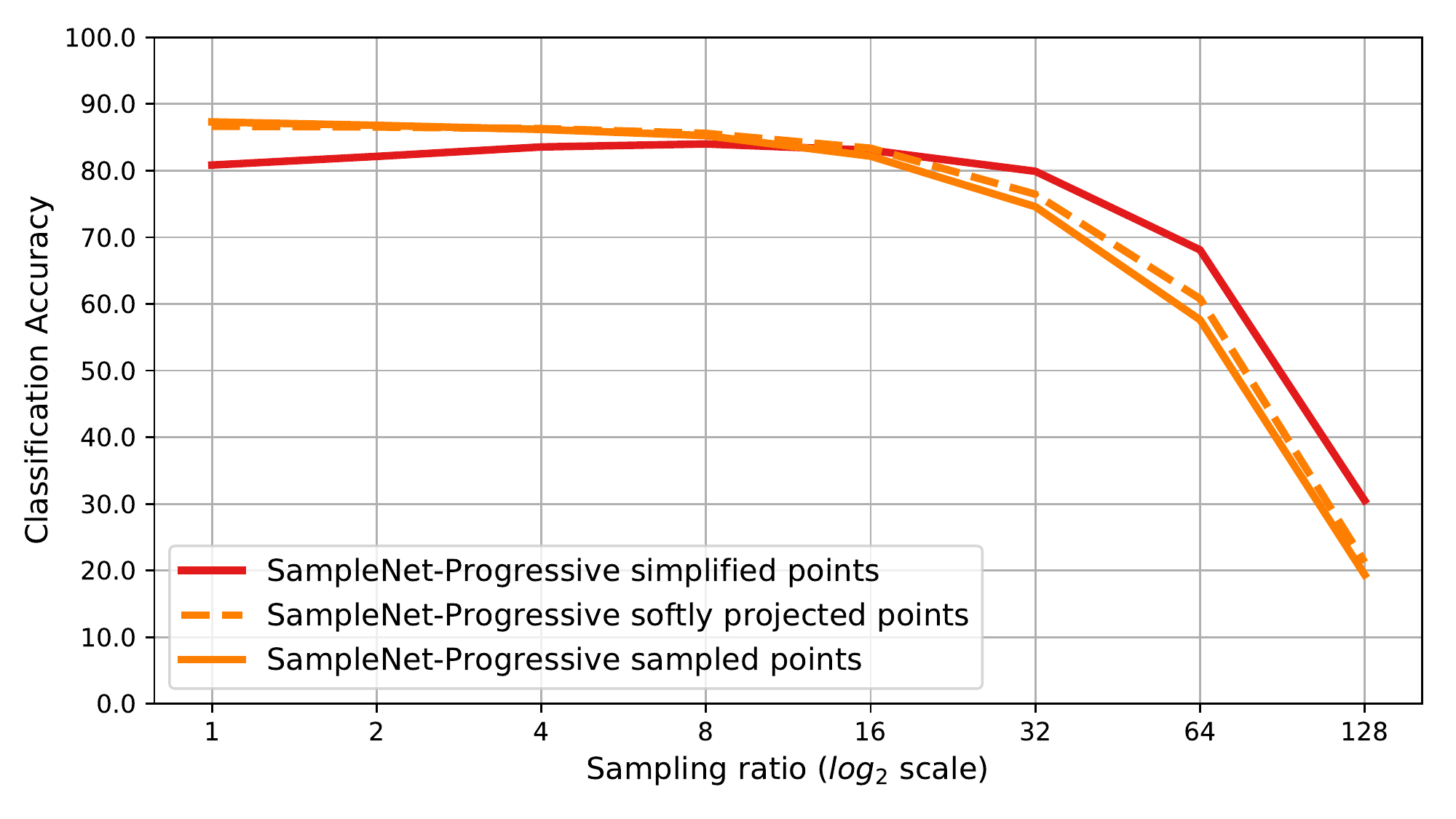}
\caption{{\bfseries Classification accuracy with simplified, softly projected, and sampled points.} The instance classification accuracy over the test set of ModelNet40 is measured with simplified, softly projected, and sampled points of \SPNetwo-Progressive. The accuracy with simplified points is either lower (up to ratio 16) or higher (from ratio 16) than that of the sampled points. On the contrary, the softly projected points closely approximate the accuracy achieved by the sampled points.}
\label{fig:points_comparison}
\end{figure}

\medskip
\noindent {\bfseries Simplified, softly projected and sampled points} \quad We evaluated the classification accuracy with simplified, softly projected, and sampled points of \SPNet for progressive sampling (denoted as \SPNetwo-Progressive). Results are reported in Figure~\ref{fig:points_comparison}. For sampling ratios up to $16$, the accuracy with simplified points is considerably lower than that of the sampled points. For higher ratios, it is the other way around. On the other hand, the accuracy with softly projected points is very close to that of the sampled ones. This indicates that our network learned to select optimal points for the task from the input point cloud, by approximating sampling with the differentiable soft projection operation.


\medskip
\noindent {\bfseries Weight evolution} \quad We examine the evolution of projection weights over time to gain insight into the behavior of the soft projection operation. We train \SPNet for $N_e \in \{1, 10, 100, 150, 200, \dots, 500\}$ epochs and apply it each time on the test set of ModelNet40. The projection weights are computed for each point and averaged over all the point clouds of the test set.

\begin{figure}[htb!]
\includegraphics[width=\columnwidth]{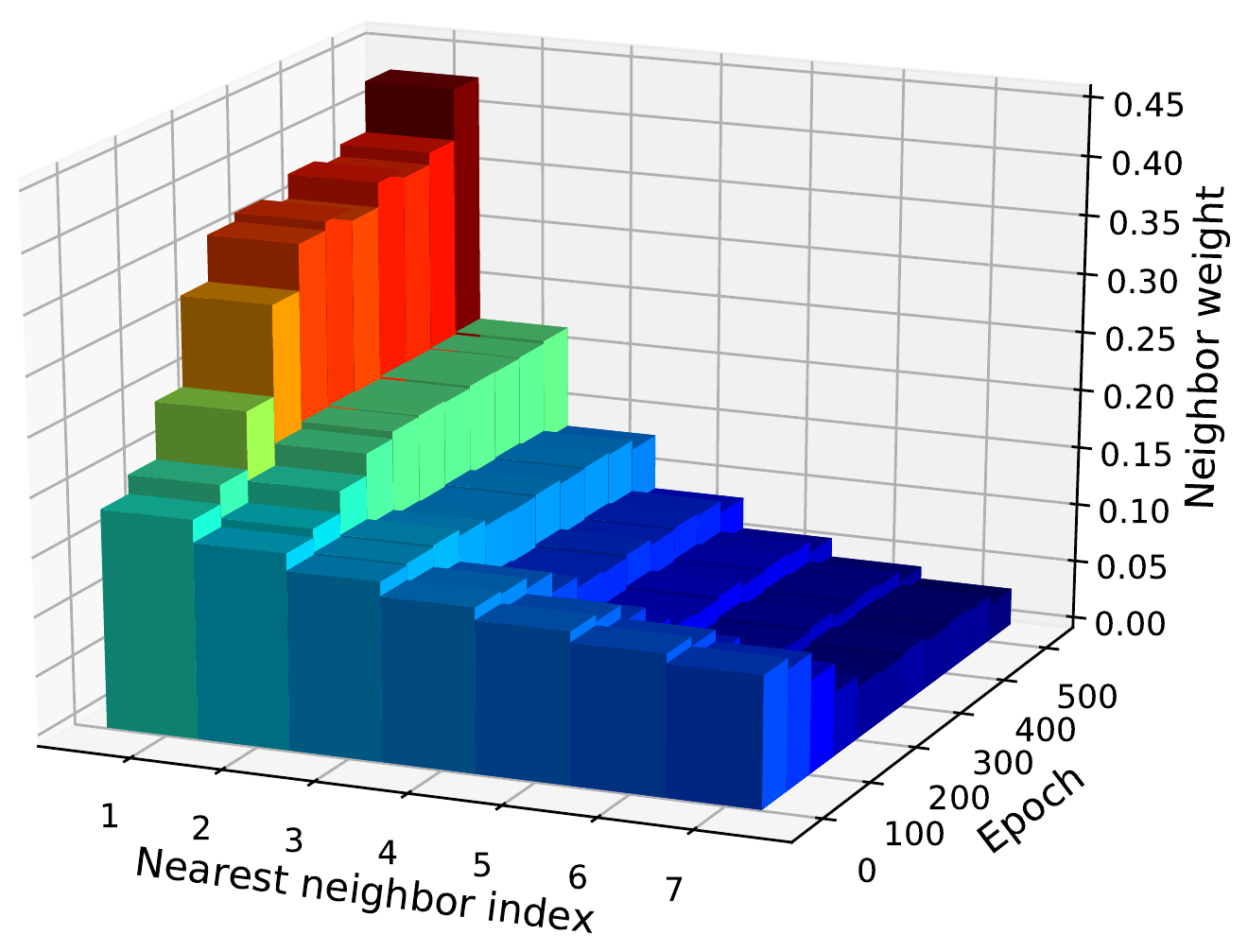}
\caption{{\bfseries Evolution of the soft projection weights.} \SPNet is trained to sample 64 points. During training, it is applied to the test split of ModelNet40. The soft projection weights are computed with $k=7$ neighbors (Equation~\ref{eq:w_i}) and averaged over all the examples of the test set. Higher bar with warmer color represents higher weight. As the training progresses, the weight distribution becomes more centered at the close neighbors.}
\label{fig:weight_evolution}
\end{figure}

Figure~\ref{fig:weight_evolution} shows the average projection weights for \SPNet trained to sample 64 points. At the first epoch, the weights are close to a uniform distribution, with a maximal and minimal weight of 0.19 and 0.11, respectively. During training, the first nearest neighbor's weight increases, while the weights of the third to the seventh neighbor decrease. The weight of the first and last neighbor converges to 0.43 and 0.03, respectively. Thus, the approximation of the nearest neighbor point by the soft projection operation is improved during training.

Interestingly, the weight distribution does not converge to a delta function at the first nearest neighbor. We recall that the goal of our learned sampling is to seek optimal points for a subsequent task. As depicted in Figure~\ref{fig:points_comparison}, similar performance is achieved with the softly projected and the sampled points. Thus, the approximation of the nearest neighbor, as done by our method, suffices.

To further investigate this subject, we trained \SPNet with additional loss term: a cross-entropy loss between the projection weight vector and a 1-hot vector, representing the nearest neighbor index. We also tried an entropy loss on the projection weights. In these cases, the weights do converge to a delta function. However, we found out that this is an over constraint, which hinders the exploration capability of \SPNetwo. Details are reported in the supplemental.



\noindent {\bfseries Temperature profile} \quad The behavior of the squared temperature coefficient ($t^2$ in Equation~\ref{eq:w_i}) during training is regarded as temperature profile. We study the influence of the temperature profile on the inference classification accuracy. Instead of using a learned profile via the projection loss in Equation~\ref{eq:loss_total_progressive}, we set $\lambda=0$ and use a pre-determined profile.

Several profiles are examined: linear rectified, exponential, and constant. The first one represents slow convergence; the exponential one simulates convergence to a lower value than that of the learned profile; the constant profile is set to $1$, as the initial temperature.

\begin{figure}[tb!]
\includegraphics[width=\columnwidth]{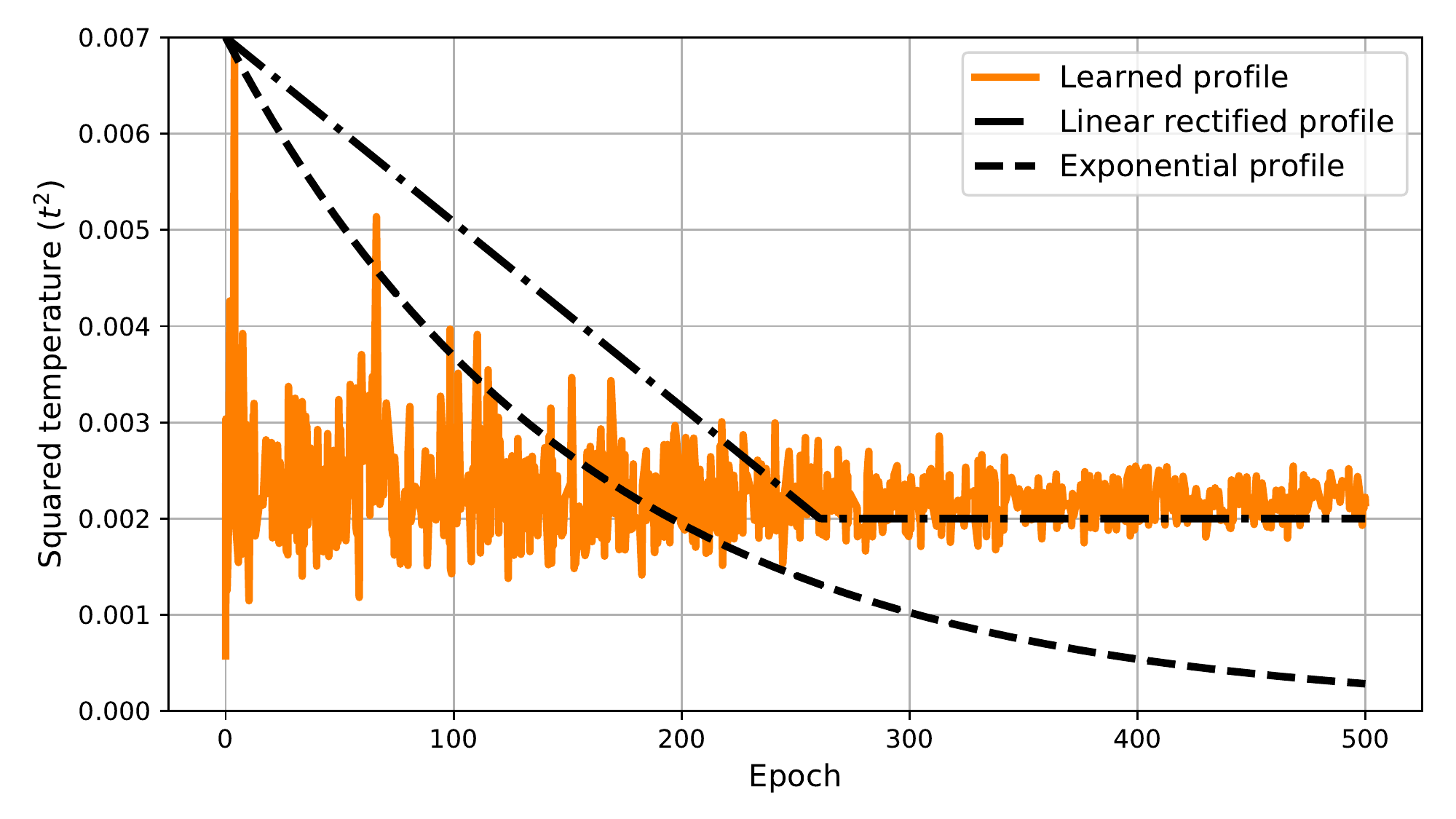}
\caption{{\bfseries Temperature profile.} Several temperature profiles are used for the training of \SPNetwo-Progressive: a learned profile; a linear rectified profile, representing slow convergence; and an exponential profile, converging to a lower value than the learned one. The classification accuracy for \SPNetwo-Progressive, trained with different profiles, is reported in Table~\ref{tbl:temperature_profile}.}
\label{fig:temperature_profile}
\end{figure}

The first two profiles and the learned profile are presented in Figure~\ref{fig:temperature_profile}. Table~\ref{tbl:temperature_profile} shows the classification accuracy with sampled points of \SPNetwo-Progressive, which was trained with different profiles. Both linear rectified and exponential profiles result in similar performance of the learned profile, with a slight advantage to the latter. However, a constant temperature causes substantial performance degradation, which is even worse than that of FPS. It indicates that a decaying profile is required for the success of \SPNetwo. Yet, is it robust to the decay behavior.

\begin{table}[tb!]
\begin{center}
\begin{tabular}{ l | c c c c c c c}
\hline
SR & 2 & 4 & 8 & 16 & 32 & 64 & 128 \\
\hline
\hline
FPS & 85.6 & 81.2 & 68.1 & 49.4 & 29.7 & 16.3 & 8.6 \\
Con & 85.5 & 75.8 & 49.6 & 32.7 & 17.1 & 7.0 & 4.7 \\
Lin & 86.7 & 86.0 & 85.0 & \bfseries 83.1 & 73.7 & 50.9 & 20.5 \\
Exp & 86.6 & 85.9 & \bfseries 85.6 & 82.0 & 74.2 & 55.6 & \bfseries 21.4 \\
Lrn & \bfseries 86.8 & \bfseries 86.2 & 85.3 & 82.2 & \bfseries 74.6 & \bfseries 57.6 & 19.4 \\
\hline
\end{tabular}
\end{center}
\caption{{\bfseries Classification accuracy with different temperature profiles.} SR stands for sampling ratio. Third to last rows correspond to \SPNetwo-Progressive trained with constant (Con), linear rectified (Lin), exponential (Exp), and learned (Lrn) temperature profile, respectively. \SPNetwo-Progressive is robust to the decay behavior of the profile. However, if the temperature remains constant, the classification accuracy degrades substantially.}
\label{tbl:temperature_profile}
\end{table}

\medskip
\noindent {\bfseries Time, space, and performance} \quad
\SPNet offers a trade-off between time, space, and performance. For example, employing \SPNet for sampling 32 points before PointNet saves about 90\% of the inference time, with respect to applying PointNet on the original point clouds. It requires only an additional 6\% memory space and results in less than 10\% drop in the classification accuracy. The computation is detailed in the supplementary.

\subsection{Registration}
We follow the work of Sarode \etal~\cite{sarode2019pcrnet} and their proposed PCRNet to construct a point cloud registration network. Point sets with 1024 points of the Car category in ModelNet40 are used. For training, we generate 4925 pairs of source and template point clouds from examples of the train set. The template is rotated by three random Euler angles in the range of \([-45\degree, 45\degree]\) to obtain the source. An additional 100 source-template pairs are generated from the test split for performance evaluation. Experiments with other shape categories appear in the supplemental.

PCRNet is trained on complete point clouds with two supervision signals: the ground truth rotation and the Chamfer distance~\cite{achlioptas2018learning} between the registered source and template point clouds. To train \SPNetwo, we freeze PCRNet and apply the same sampler to both the source and template. The registration performance is measured in mean rotation error (MRE) between the estimated and the ground truth rotation in angle-axis representation. More details regarding the loss terms and the evaluation metric are given in the supplementary material.

\begin{figure}[htb!]
\includegraphics[width=\columnwidth]{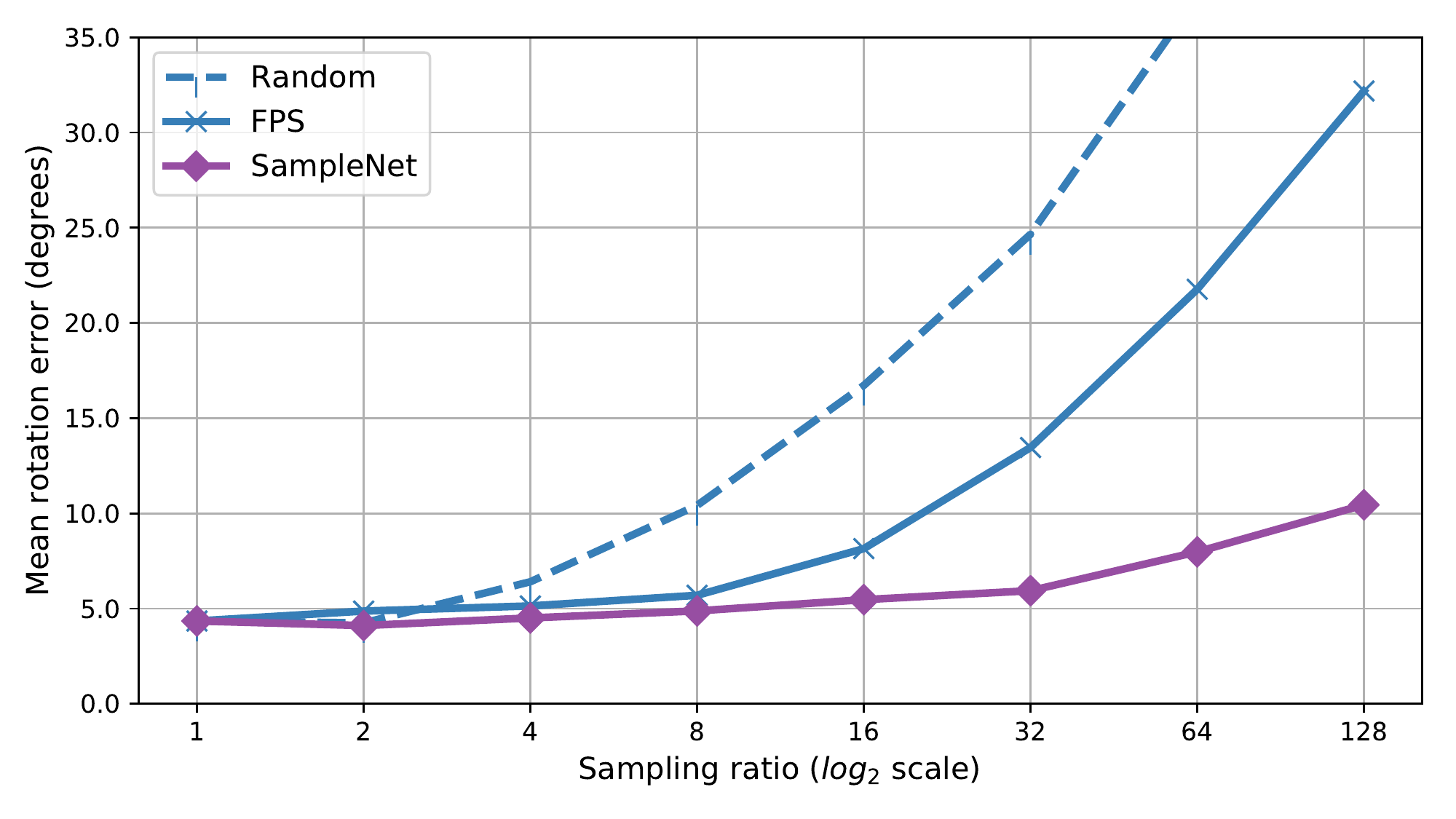}
\caption{{\bfseries Rotation error with \SPNetwo.} PCRNet is used as the task network for registration. It was trained on complete point clouds of 1024 points from the Car category in ModelNet40. Mean rotation error (MRE) between registered source and template point cloud pairs is measured on the test split for different sampling methods. Our \SPNet achieves the lowest MRE for all ratios.}
\label{fig:spnet_registration}
\end{figure}

The sampling method of Dovrat \etal~\cite{dovrat2019learning} was not applied for the registration task, and much work is needed for its adaption. Thus, for this application, we utilize FPS and random sampling as baselines. Figure~\ref{fig:spnet_registration} presents the MRE for different sampling methods. The MRE with our proposed sampling remains low, while for the other methods, it is increased with the sampling ratio. For example, for a ratio of 32, the MRE with \SPNet is $5.94\degree$, while FPS results in a MRE of $13.46\degree$, more than twice than \SPNetwo.


\begin{figure}[htb!]
\begin{center}
\begin{tabular}{c c}
\includegraphics[width=0.45\linewidth]{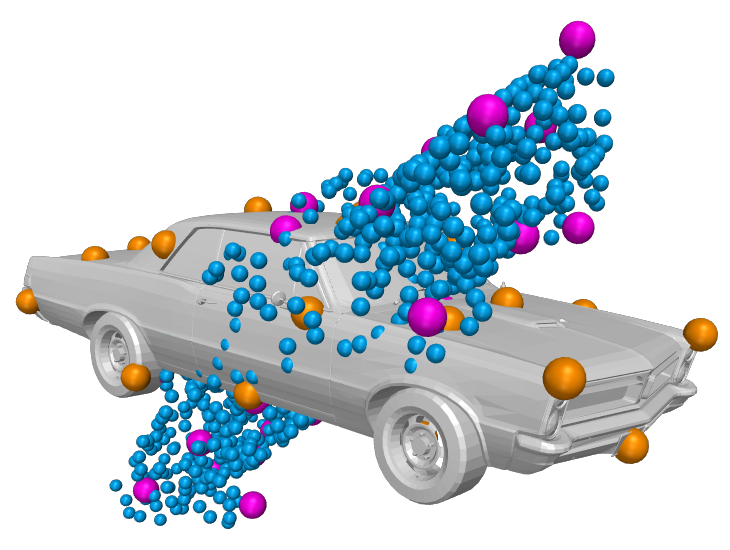} &
\includegraphics[width=0.45\linewidth]{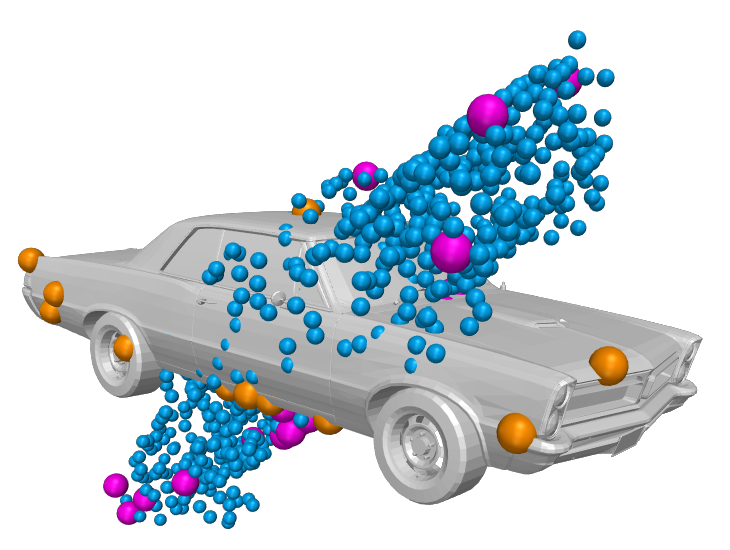} \\
\includegraphics[width=0.45\linewidth]{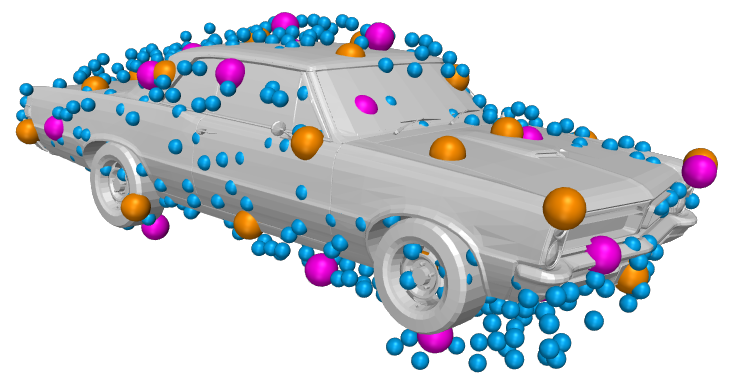} & \includegraphics[width=0.45\linewidth]{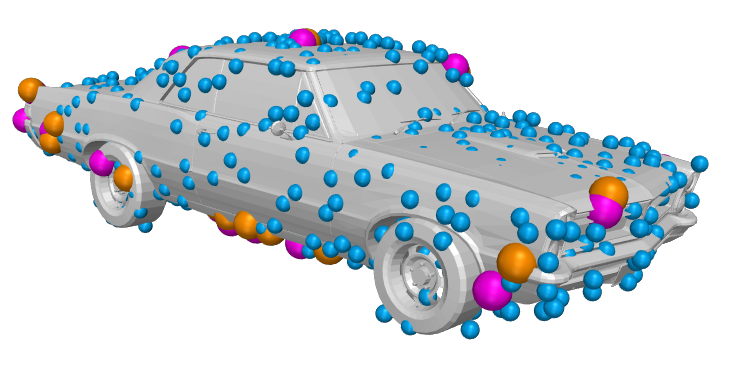} \\
FPS & \SPNet \\
\end{tabular}
\caption{{\bfseries Registration with sampled points.} Top row: unregistered source with 1024 points in Blue overlaid on the mesh model. Sampled sets of 32 points from the template and source are illustrated in Orange and Magenta, respectively. Bottom row: the registered source cloud is overlaid on the mesh. \SPNet enables us to perform registration of point clouds from their samples.}
\label{fig:reg_car}
\end{center}
\end{figure}

A registration example is visualized in Figure~\ref{fig:reg_car}. FPS points are taken uniformly, while \SPNet points are located at semantic features of the shape. Using FPS does not enable to align the sampled points, as they are sampled at different parts of the original point cloud. In contrast, \SPNet learns to sample similar points from different source and template clouds. Thus, registration with its sampled sets is possible. Quantitative measure of this sampling consistency is presented in the supplementary.




In conclusion, \SPNet proves to be an efficient sampling method for the registration task, overcoming the challenge of sampling two different point clouds. We attribute this success to the permutation invariance of \SPNetwo, as opposed to FPS and random sampling. That, together with the task-specific optimization, gives \SPNet the ability to achieve low registration error.


\subsection{Reconstruction}
\SPNet is applied to the reconstruction of points clouds from sampled points. The task network, in this case, is the autoencoder of Achlioptas \etal~\cite{achlioptas2018learning} that was trained on point clouds with 2048 points. The sampling ratio is defined as $2048/m$, where $m$ is the sample size.

We evaluate the reconstruction performance by normalized reconstruction error (NRE)~\cite{dovrat2019learning}. The reconstruction error is the Chamfer distance~\cite{achlioptas2018learning} between a reconstructed point cloud and the complete input set. The NRE is the error when reconstructing from a sampled set divided by the error of reconstruction from the complete input.

\begin{figure}[tb!]
\includegraphics[width=\columnwidth]{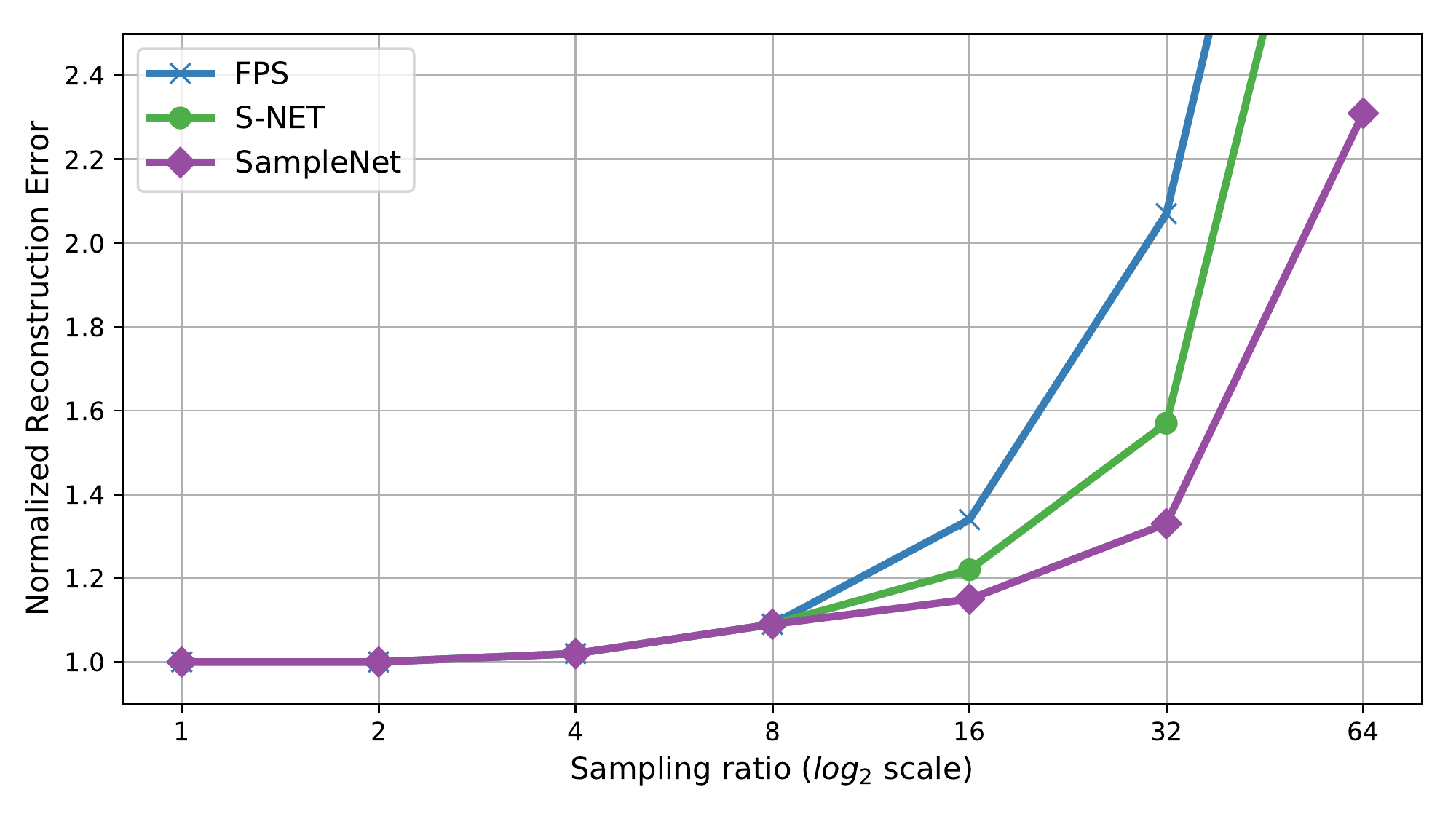}
\caption{{\bfseries \SPNet for reconstruction.} The input point cloud is reconstructed from its sampled points. The reconstruction error is normalized by the error when using the complete input point set. Starting from ratio $8$, \SPNet achieves lower error, with an increasing gap in the sampling ratio.}
\label{fig:SPNet_reconstruction}
\end{figure}

Figure~\ref{fig:SPNet_reconstruction} reports the average NRE for the test split of the shape classes we use from ShapeNet database. Up to sampling ratio of $8$, all the methods result in similar reconstruction performance. However, for higher ratios, \SPNet outperforms the other alternatives, with an increasing margin. For example, for a sampling ratio of 32, the NRE for S-NET is 1.57 versus 1.33 for \SPNet - a reduction of 24\%. We conclude that \SPNet learns to sample useful points for reconstructing point sets unseen during training.

\begin{figure}[ht!]
\begin{center}
\begin{tabular}{ c c c c }
\includegraphics[width=0.18\linewidth]{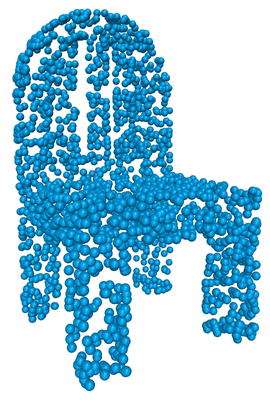} &
\includegraphics[width=0.18\linewidth]{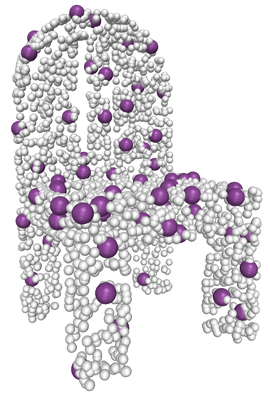} &
\includegraphics[width=0.18\linewidth]{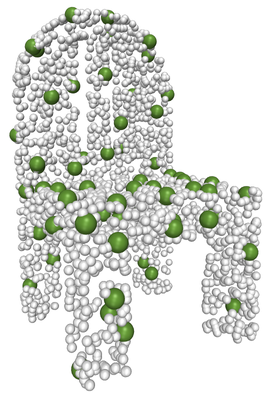} &
\includegraphics[width=0.18\linewidth]{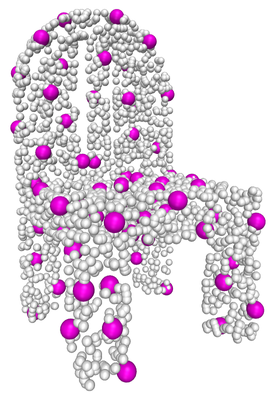} \\
\includegraphics[width=0.18\linewidth]{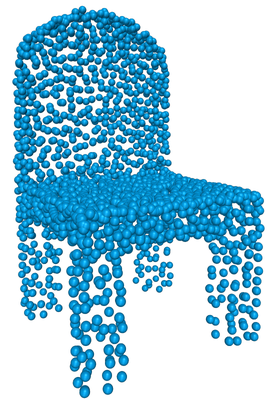} &
\includegraphics[width=0.18\linewidth]{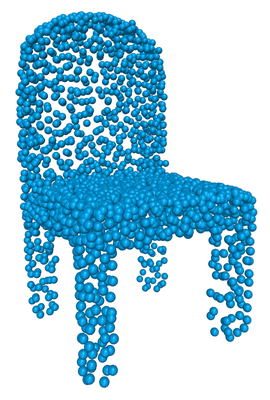} &
\includegraphics[width=0.18\linewidth]{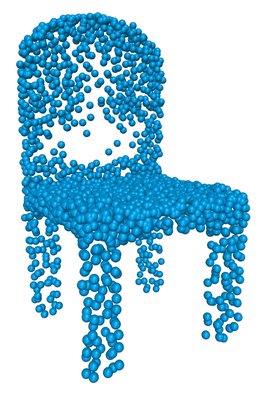} &
\includegraphics[width=0.18\linewidth]{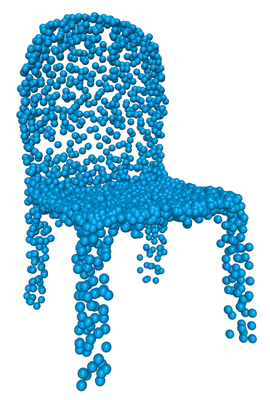} \\
\includegraphics[width=0.20\linewidth]{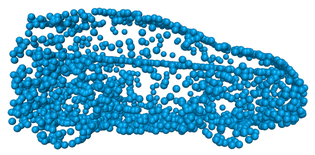} &
\includegraphics[width=0.20\linewidth]{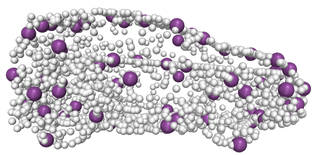} &
\includegraphics[width=0.20\linewidth]{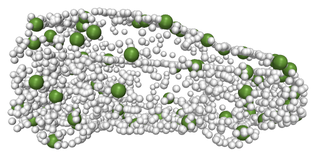} &
\includegraphics[width=0.20\linewidth]{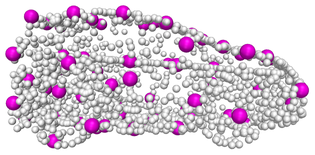} \\
\includegraphics[width=0.20\linewidth]{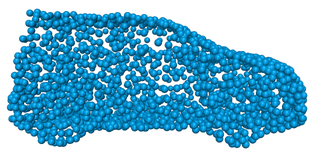} &
\includegraphics[width=0.20\linewidth]{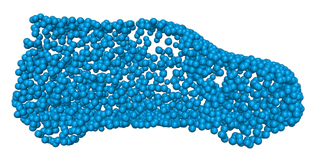} &
\includegraphics[width=0.20\linewidth]{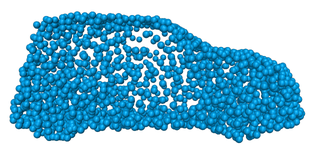} &
\includegraphics[width=0.20\linewidth]{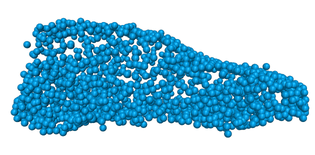} \\
Complete & \SPNet & S-NET & FPS \\
\end{tabular}
\caption{{\bfseries Reconstruction from sampled points.} Top and third rows: complete input point set of 2048 points, input with 64 \SPNet points (in Purple), input with 64 S-NET points (in Green), input with 64 FPS points (in Magenta). Second and bottom rows: reconstructed point cloud from the input and the corresponding sample. Using \SPNet better preserves the input shape and results in similar reconstruction to one from the complete input.}
\label{fig:rcon_vis}
\end{center}
\end{figure}

Reconstruction from samples is visualized in Figure~\ref{fig:rcon_vis}. FPS points are spread over the shape uniformly, as opposed to the non-uniform pattern of \SPNet and S-NET. Interestingly, some points of the learned sampling methods are sampled in similar locations, for example, at the legs of the chair. Nevertheless, reconstructing using S-NET or FPS points results in artifacts or loss of details. On the contrary, utilizing \SPNet reconstructs the input shape better.

\begin{figure}[htb!]
\begin{center}
\begin{tabular}{ c c c }
\includegraphics[width=0.24\linewidth]{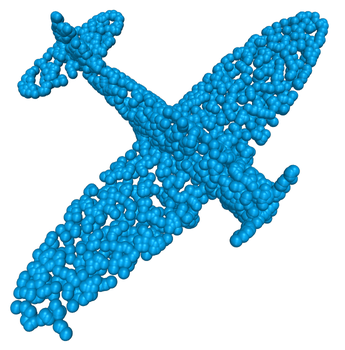} &
\includegraphics[width=0.24\linewidth]{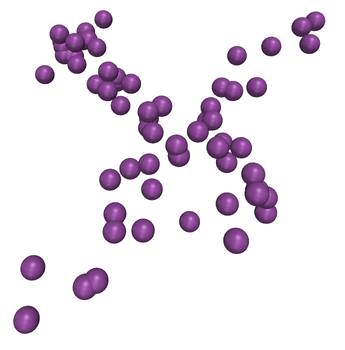} &
\includegraphics[width=0.24\linewidth]{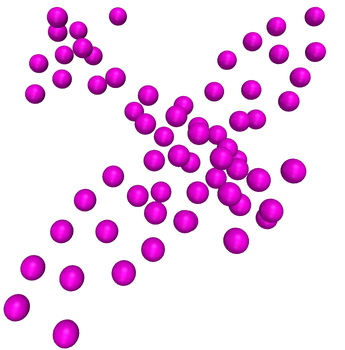} \\
\includegraphics[width=0.24\linewidth]{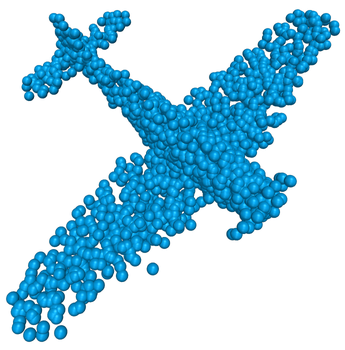} &
\includegraphics[width=0.24\linewidth]{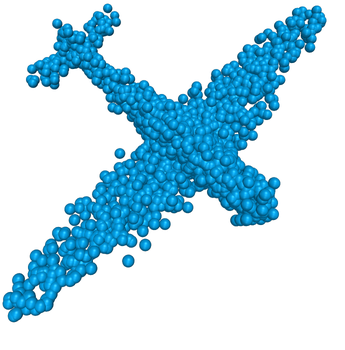} &
\includegraphics[width=0.24\linewidth]{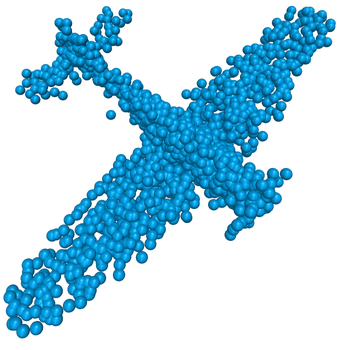} \\
Complete & \SPNet & FPS \\
\end{tabular}
\caption{{\bfseries A failure example.} Top row: complete input with 2048 points, 64 \SPNet points (in Purple), 64 FPS points (in Magenta). Bottom row: reconstruction from complete input and from corresponding sampled points. In this case, uniform sampling by FPS is preferred.}
\label{fig:failure_vis}
\end{center}
\end{figure}

\medskip
\noindent {\bfseries A failure case} \quad
When computing the NRE per shape class, \SPNet achieves lower NRE for Chair, Car, and Table classes. However, the NRE of FPS is better than that of \SPNet for airplanes. For example, for a sample size of 64 points, the NRE of FPS is 1.31, while the NREs of \SPNet and S-NET are 1.39 and 1.41, respectively. Figure~\ref{fig:failure_vis} shows an example of reconstructing an airplane from 64 points. FPS samples more points on the wings than \SPNetwo. These points are important for the reconstruction of the input, thus leading to an improved result.

\section{Conclusions} \label{sec:conclusions}

We presented a learned sampling approach for point clouds. Our network, \SPNetwo, takes an input point cloud and produces a smaller point cloud that is optimized to some downstream task. The key challenge was to deal with the non-differentiability of the sampling operation. To solve this problem, we proposed a differentiable relaxation, termed soft projection, that represents output points as a weighted average of points in the input. During training, the projection weights were optimized to approximate nearest neighbor sampling, which occurs at the inference phase. The soft projection operation replaced the regression of optimal points in the ambient space with multiple classification problems in local neighborhoods of the input.

We applied our technique to point cloud classification and reconstruction. We also evaluated our method on the task of point cloud registration. The latter is more challenging than previous tasks because it requires the sampling to be consistent across two different point clouds. We found that our method consistently outperforms the competing non-learned as well as learned sampling alternatives by a large margin.

\medskip
\noindent {\bfseries Acknowledgment} \quad
This work was partly funded by ISF grant number 1549/19.




{\small
\bibliographystyle{ieee_fullname}
\bibliography{references}
}

\clearpage

\appendix
\section*{Supplementary}

In the following sections, we provide additional details and results of our sampling approach. Section~\ref{sec:supp_additional_results} presents additional results of our method. An ablation study is reported in Section~\ref{sec:supp_ablation_study}. Section~\ref{sec:supp_mathematical_aspects} describes mathematical aspects of the soft projection operation, employed by \SPNetwo. Finally, experimental settings, including network architecture and hyperparameter settings, are given in Section~\ref{sec:supp_experimental_settings}.

\section{Additional results} \label{sec:supp_additional_results}

\subsection{Point cloud retrieval}
We employ sampled point sets for point cloud retrieval, using either farthest point sampling (FPS), S-NET, or \SPNetwo. In order to evaluate cross-task usability, the last two sampling methods are trained with PointNet for classification and applied for the retrieval task without retraining~\cite{dovrat2019learning}. The shape descriptor is the activation vector of the second-last layer of PointNet when it fed with sampled or complete clouds. The distance metric is $l_2$ between shape descriptors.

\begin{figure}[htb!]
\includegraphics[width=\columnwidth]{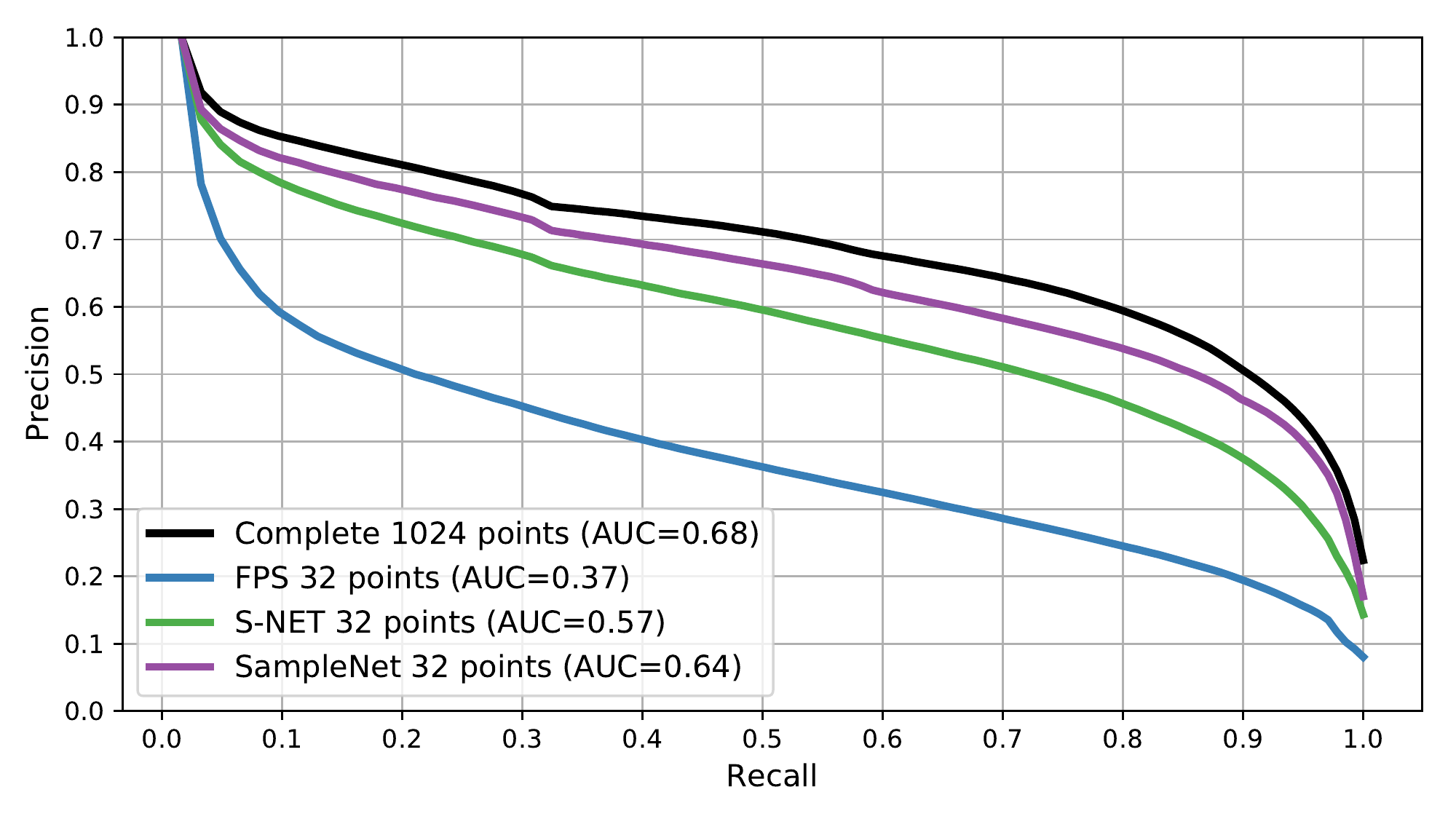}
\caption{{\bfseries Precision-recall curve with sampled points.} PointNet is fed with sampled point clouds from the test set. Its penultimate layer is used as the shape descriptor. Utilizing \SPNet results in improved retrieval performance in comparison to the other sampling methods. Using only 32 points, \SPNet is close to the precision obtained with complete input points cloud, with a drop of only 4\% in the area under the curve (AUC).}
\label{fig:precision_recall}
\end{figure}

Precision and recall are evaluated on the test set of ModelNet40, where each shape is used as a query. The results when using the complete 1024 point sets and samples of 32 points are presented in Figure~\ref{fig:precision_recall}. \SPNet improves the precision over all the recall range with respect to S-NET and approaches the performance with complete input sets. It shows that the points sampled by \SPNet are suitable not only for point cloud classification but also for retrieval.

\subsection{Progressive sampling}
\begin{figure}[htb!]
\includegraphics[width=\columnwidth]{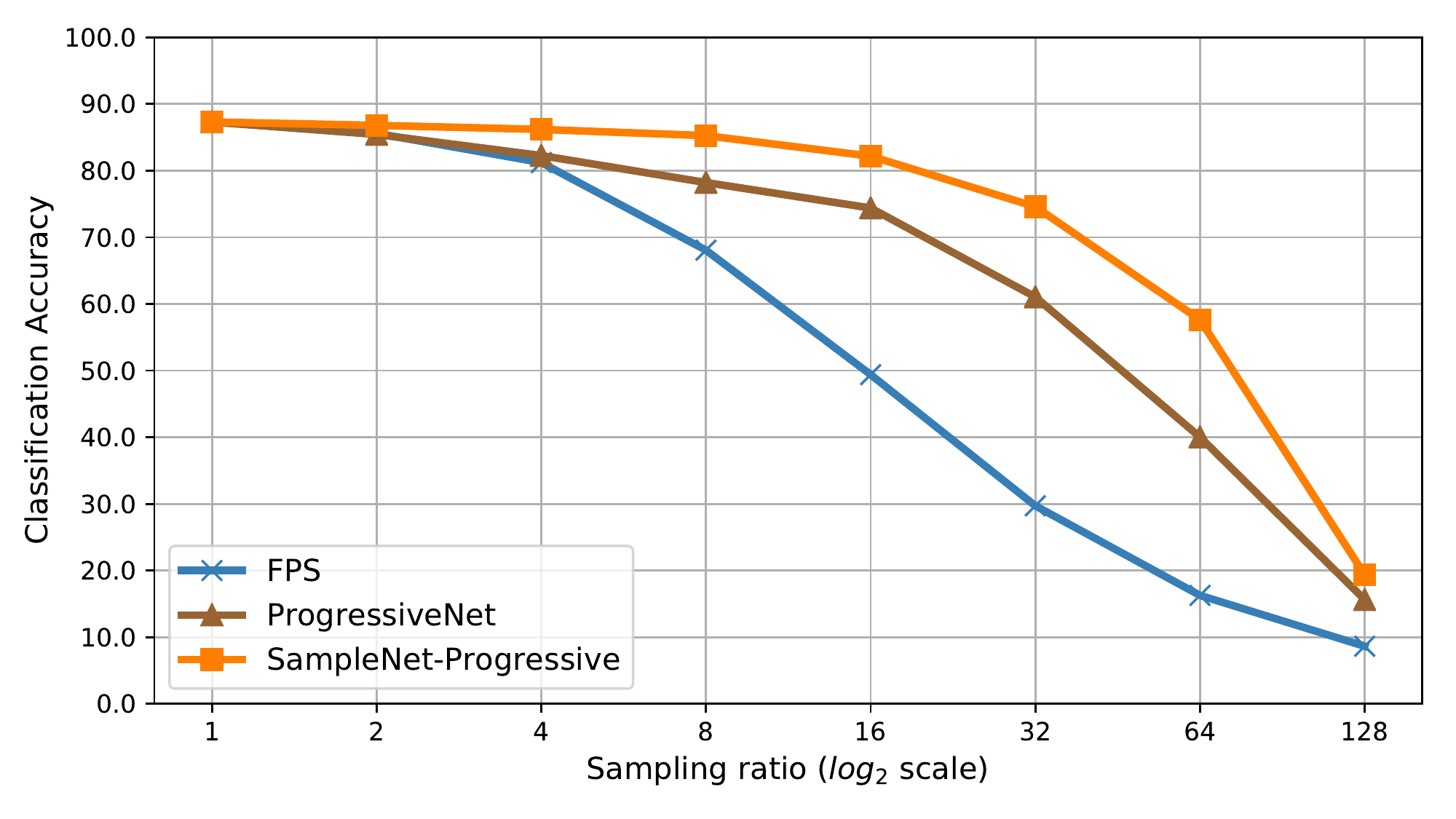}
\caption{{\bfseries Classification results with \SPNetwo-Progressive.} PointNet vanilla is used as the task network and was pre-trained on point clouds with 1024 points. The instance classification accuracy is evaluated on sampled point clouds from the test split. Our sampling network outperforms farthest point sampling (FPS) and ProgressiveNet~\cite{dovrat2019learning}.}
\label{fig:SPNet_prog_classification}
\end{figure}

Our method is applied to the progressive sampling of point clouds~\cite{dovrat2019learning} for the classification task. In this case, the vanilla version of PointNet~\cite{qi2017pointnet} is employed as the classifier~\cite{dovrat2019learning}. Performance gains are achieved in the progressive sampling settings, as shown in Figure~\ref{fig:SPNet_prog_classification}. They are smaller than those of \SPNet trained per sample size separately (see Figure~\ref{fig:SPNet_single_classification} in the main body) since for progressive sampling, \SPNetwo-Progressive should be optimal for all the control sizes concurrently.

\begin{figure}[htb!]
\includegraphics[width=\columnwidth]{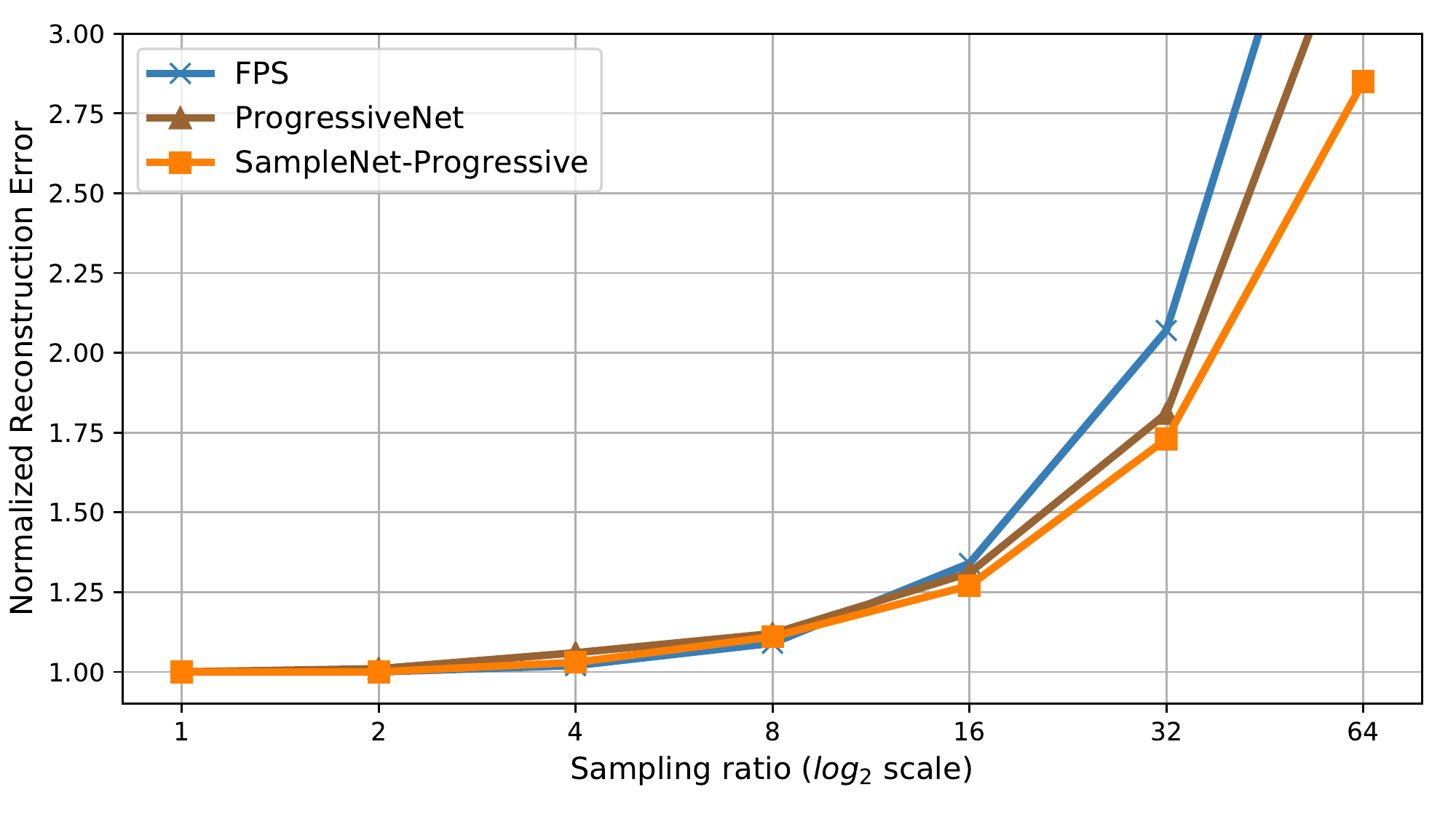}
\caption{{\bfseries Normalized reconstruction error with \SPNetwo-Progressive.} Point clouds are reconstructed from nested sets of sampled points. We normalize the reconstruction error from a sample by the error resulting from a complete input. As the sampling ratio is increased, the improvement of \SPNetwo, compared to the alternatives, becomes more dominant.}
\label{fig:SPNet_prog_reconstruction}
\end{figure}

\begin{figure*}[tb!]
\begin{center}
\begin{tabular}{ c | c c c c }

             & \SPNetwo- & \SPNetwo- & \SPNetwo- & \SPNetwo- \\
Input $2048$ & Progressive $32$ & Progressive $64$ & Progressive $128$ & Progressive $256$ \\

\includegraphics[width=0.17\linewidth]{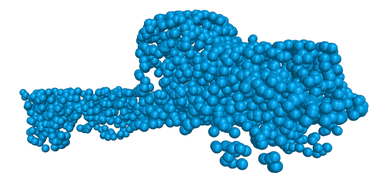} &
\includegraphics[width=0.17\linewidth]{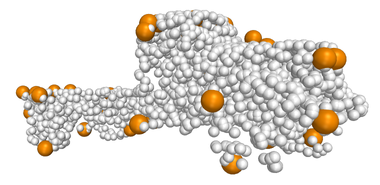} &
\includegraphics[width=0.17\linewidth]{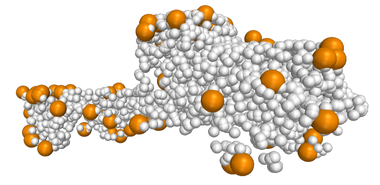} &
\includegraphics[width=0.17\linewidth]{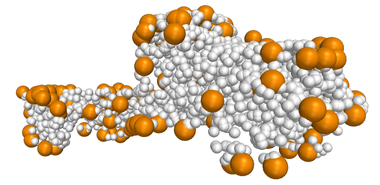} &
\includegraphics[width=0.17\linewidth]{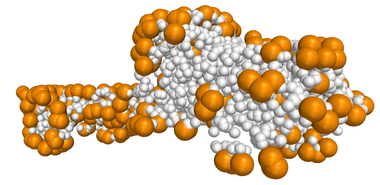} \\

Reconstruction & \multicolumn{4}{c}{Reconstructions from \SPNetwo-Progressive samples}  \\
\includegraphics[width=0.17\linewidth]{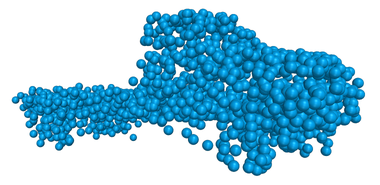} &
\includegraphics[width=0.17\linewidth]{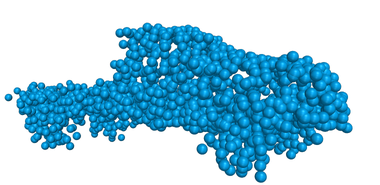} &
\includegraphics[width=0.17\linewidth]{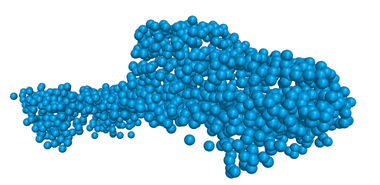} &
\includegraphics[width=0.17\linewidth]{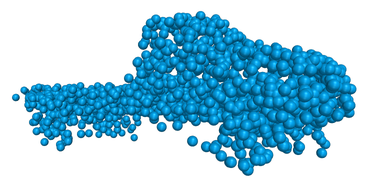} &
\includegraphics[width=0.17\linewidth]{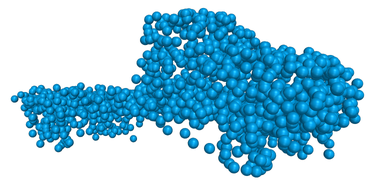} \\

\hline

Input $2048$ & ProgressiveNet $32$ & ProgressiveNet $64$ & ProgressiveNet $128$ & ProgressiveNet $256$ \\
\includegraphics[width=0.17\linewidth]{supplementary/figures/progressive_samp_k3d/car_578_input_pc.png} &
\includegraphics[width=0.17\linewidth]{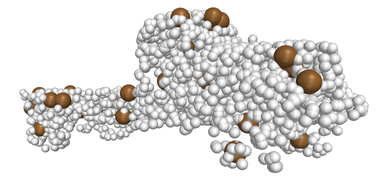} &
\includegraphics[width=0.17\linewidth]{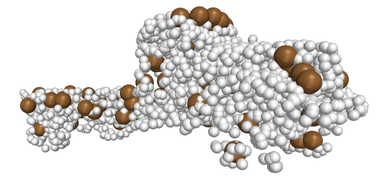} &
\includegraphics[width=0.17\linewidth]{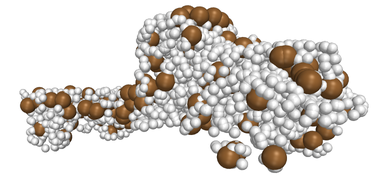} &
\includegraphics[width=0.17\linewidth]{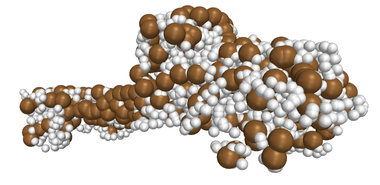} \\

Reconstruction & \multicolumn{4}{c}{Reconstructions from ProgressiveNet samples}  \\
\includegraphics[width=0.17\linewidth]{supplementary/figures/progressive_samp_k3d/car_578_rcon_input.png} &
\includegraphics[width=0.17\linewidth]{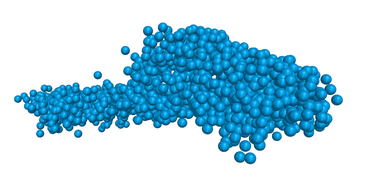} &
\includegraphics[width=0.17\linewidth]{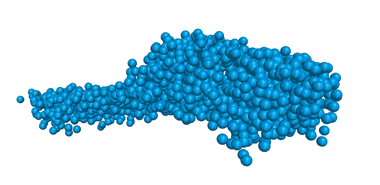} &
\includegraphics[width=0.17\linewidth]{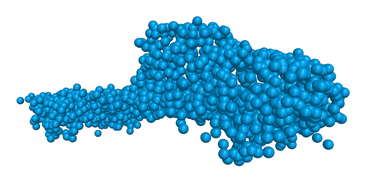} &
\includegraphics[width=0.17\linewidth]{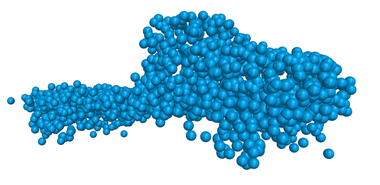} \\

\hline

Input $2048$ & FPS $32$ & FPS $64$ & FPS $128$ & FPS $256$ \\
\includegraphics[width=0.17\linewidth]{supplementary/figures/progressive_samp_k3d/car_578_input_pc.png} &
\includegraphics[width=0.17\linewidth]{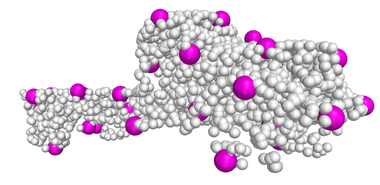} &
\includegraphics[width=0.17\linewidth]{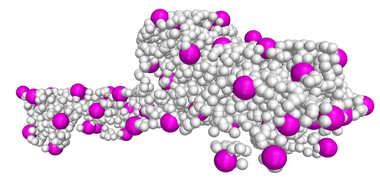} &
\includegraphics[width=0.17\linewidth]{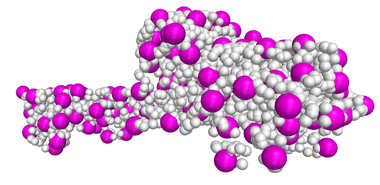} &
\includegraphics[width=0.17\linewidth]{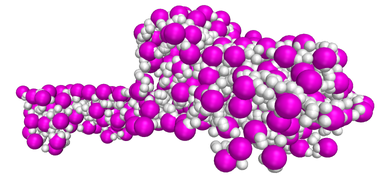} \\

Reconstruction & \multicolumn{4}{c}{Reconstructions from FPS samples}  \\
\includegraphics[width=0.17\linewidth]{supplementary/figures/progressive_samp_k3d/car_578_rcon_input.png} &
\includegraphics[width=0.17\linewidth]{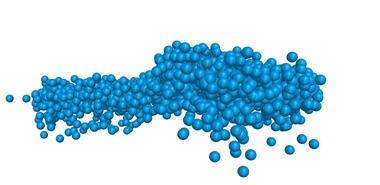} &
\includegraphics[width=0.17\linewidth]{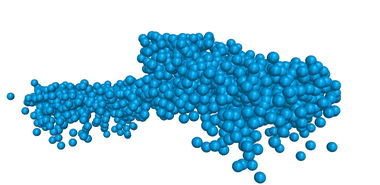} &
\includegraphics[width=0.17\linewidth]{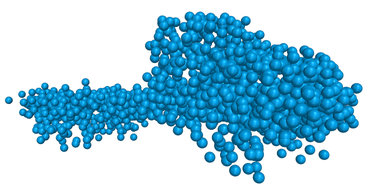} &
\includegraphics[width=0.17\linewidth]{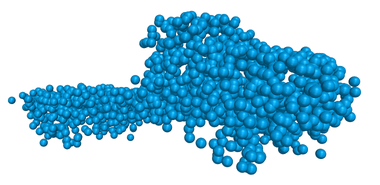} \\

\end{tabular}
\caption{{\bfseries Reconstructions with \SPNetwo-Progressive.} Odd rows: input point cloud and samples of different progressive sampling methods. The number of sampled points is denoted next to the method's name. Even rows: reconstruction from the input and the corresponding sample. Our \SPNetwo-Progressive selects most of its points at the outline of the shape, while ProgressiveNet~\cite{dovrat2019learning} selects interior points and FPS points are spread uniformly. In contrast to the other methods, our result starts to resemble the reconstruction from the complete input when using only 32 points, which is about 1.5\% of the input data.}
\label{fig:progressive_sampling}
\end{center}
\end{figure*}

We also perform reconstruction from progressively sampled point clouds. Our normalized reconstruction error is compared to that of FPS and ProgressiveNet~\cite{dovrat2019learning} in Figure~\ref{fig:SPNet_prog_reconstruction}. Figure~\ref{fig:progressive_sampling} shows a visual reconstruction example.

\subsection{Computation load and memory space}
The computation load of processing a point cloud through a network is regarded as the number of multiply-accumulate operations (MACs) for inference. The required memory space is the number of learnable parameters of the network.

For a PointNet like architecture, the number of MACs is mainly determined by the number of input points processed by the multi-layer perceptrons (MLPs). Thus, reducing the number of points reduces the computational load. The memory space of \SPNet depends on the number of output points, resulting from the last fully connected layer. The soft projection operation adds only one learnable parameter, which is negligible to the number of weights of \SPNetwo.

We evaluate the computation load and memory space for the classification application. We denote the computation and memory of \SPNet that outputs $m$ points as $C_{SN_m}$ and $M_{SN_m}$, respectively. Similarly, the computation of PointNet that operates on $m$ points is denoted as $C_{PN_m}$, and for a complete point cloud as $C_{PN}$. The memory of PointNet is marked $M_{PN}$. It is independent of the number of processed points. When concatenating \SPNet with PointNet, we define the computation reduction percent $CR$ as:
\begin{equation} \label{computation_reduction}
CR = 100 \cdot \left( 1 - \frac{C_{SN_m} + C_{PN_m}}{C_{PN}} \right),
\end{equation}

\noindent and the memory increase percent $MI$ as:
\begin{equation} \label{memory_increase}
MI = 100 \cdot \frac{M_{SN_m} + M_{PN}}{M_{PN}}.
\end{equation}

\begin{figure}[bt!]
\includegraphics[width=\columnwidth]{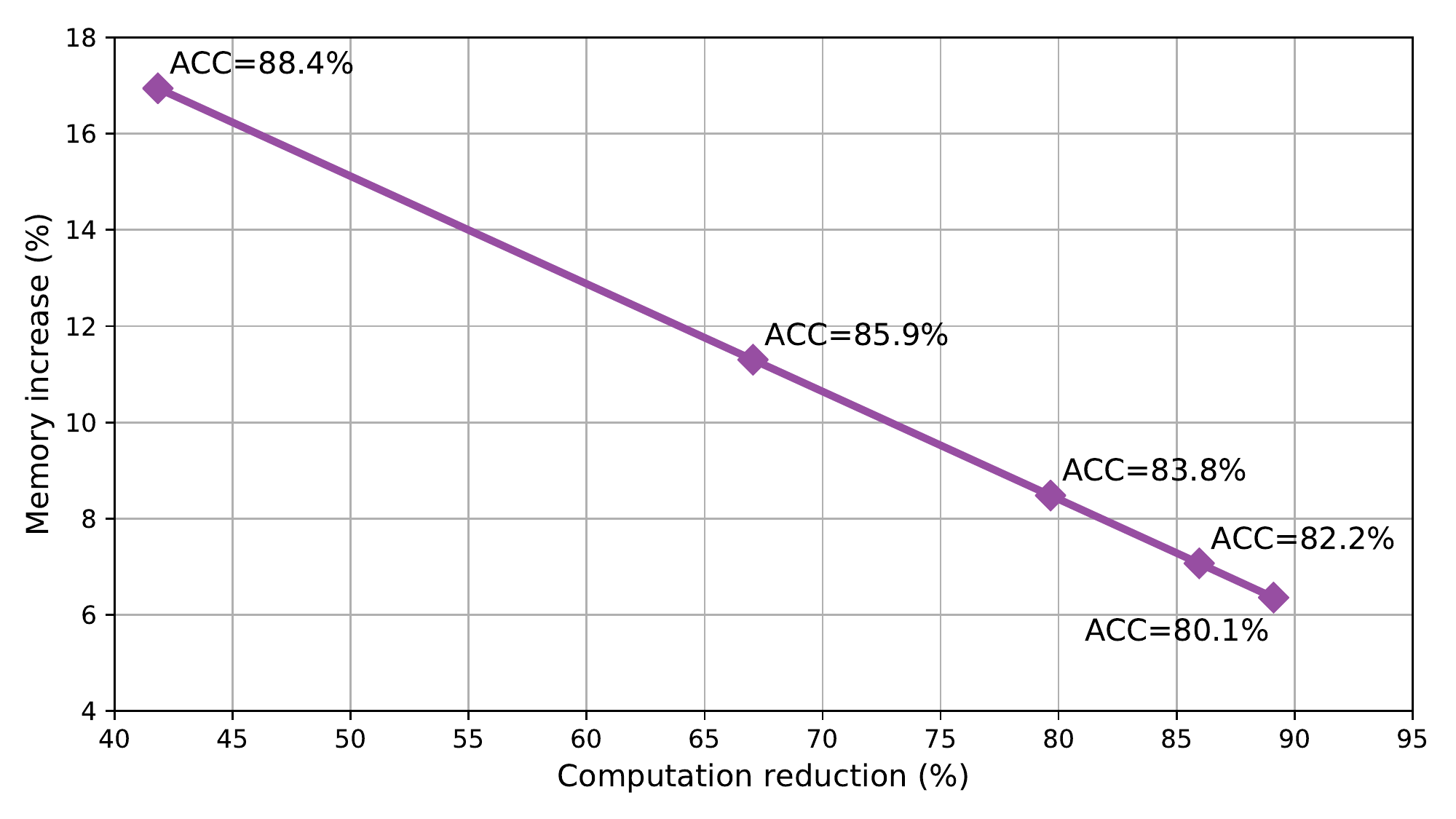}
\caption{{\bfseries Memory, computation, and performance.} The memory increase for chaining \SPNet with PointNet is plotted against the computation reduction, which results from processing sampled instead of complete clouds. The points on the graph from left to right correspond to sampling ratios $\{2, 4, 8, 16, 32\}$. ACC is the classification accuracy on the test split of ModelNet40 when PointNet runs on sampled point sets. With a slight increase in memory and small accuracy drop, \SPNet reduces the computational load substantially.}
\label{fig:time_and_space}
\end{figure}

Figure~\ref{fig:time_and_space} presents the memory increase versus computation reduction. As the number of sampled points is reduced, the memory increase is lower, and the computation reduction is higher, with a mild decrease in the classification accuracy.





For example, \SPNet for 32 points has 0.22M parameters and performs 34M MACs ('M' stands for Million). PointNet that operates on point clouds of 32 instead of 1024 points requires only 14M instead of 440M MACs. The number of PointNet parameters is 3.5M. \SPNet followed by PointNet sums up to 48M MACs and 3.72M parameters. These settings require about 6\% additional memory space and reduce the computational load by almost 90\%.



\subsection{Sampling consistency for registration task}
Given a sampled set \(T_s^{gt}\) of template points, rotated by the ground truth rotation \(R_{gt}\), and a sampled set \(S_s\) of source points, the sampling consistency is defined as the Chamfer distance between these two sets:
\begin{equation} \label{eq:reg_consistency}
\begin{split}
C(S_s,T_s^{gt}) = \frac{1}{|S_s|}\sum_{\mathbf{t} \in S_s}{\min_{\mathbf{t} \in T^{gt}_s}||\mathbf{s}-\mathbf{t}||_2^2} \\
+ \frac{1}{|T^{gt}_s|}\sum_{\mathbf{t} \in T^{gt}_s}{\min_{\mathbf{s} \in S_s}||\mathbf{t}-\mathbf{s}||_2^2}.
\end{split}
\end{equation}

For a given sampler, this metric quantifies the tendency of the algorithm to sample similar points from the source and template point clouds. We measure the average consistency on the test set of the Car category from ModelNet40. Results for random sampling, FPS and \SPNet are reported in Table~\ref{tbl:consistency}. The table shows that \SPNet sampling is substantially more consistent than that of the alternatives. This behavior can explain its success for the registration task.
 

\subsection{Registration for different shape categories}
\begin{figure}[tbh!]
\begin{center}
\begin{tabular}{c c c}
Input & FPS & \SPNet \\
\includegraphics[width=0.28\linewidth]{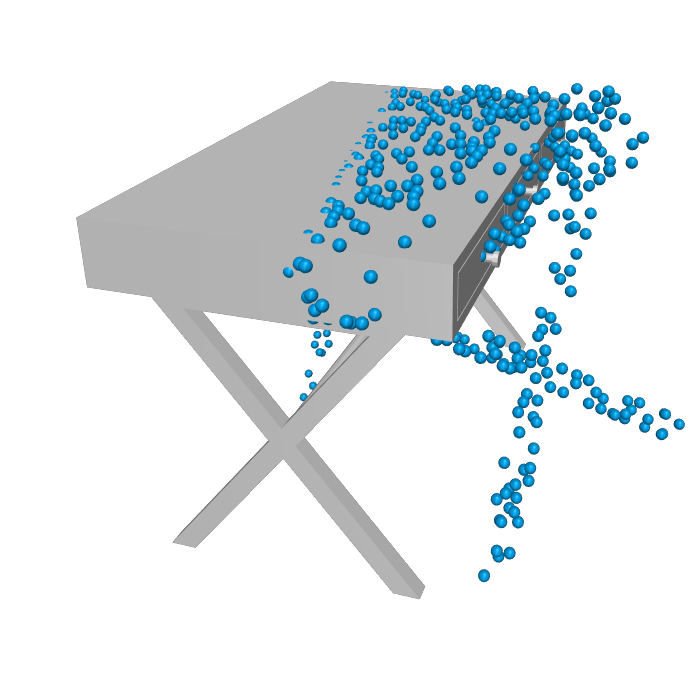} &
\includegraphics[width=0.28\linewidth]{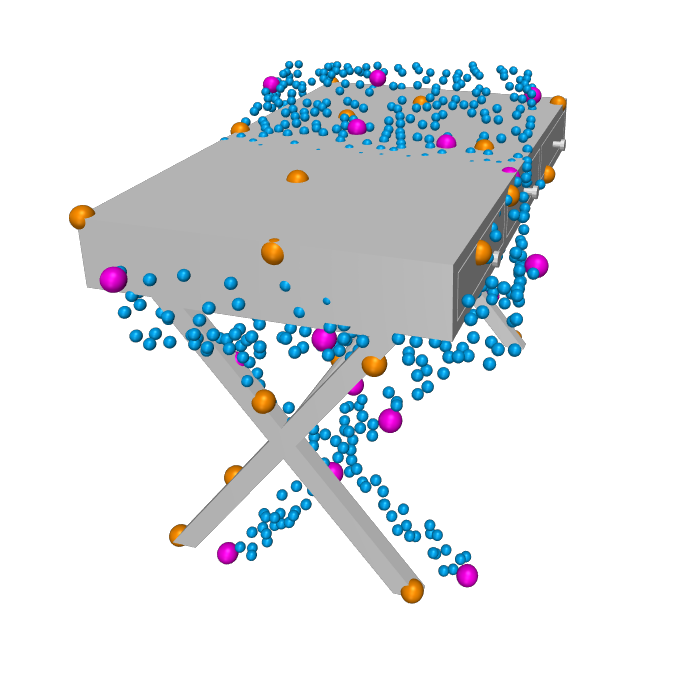} &
\includegraphics[width=0.28\linewidth]{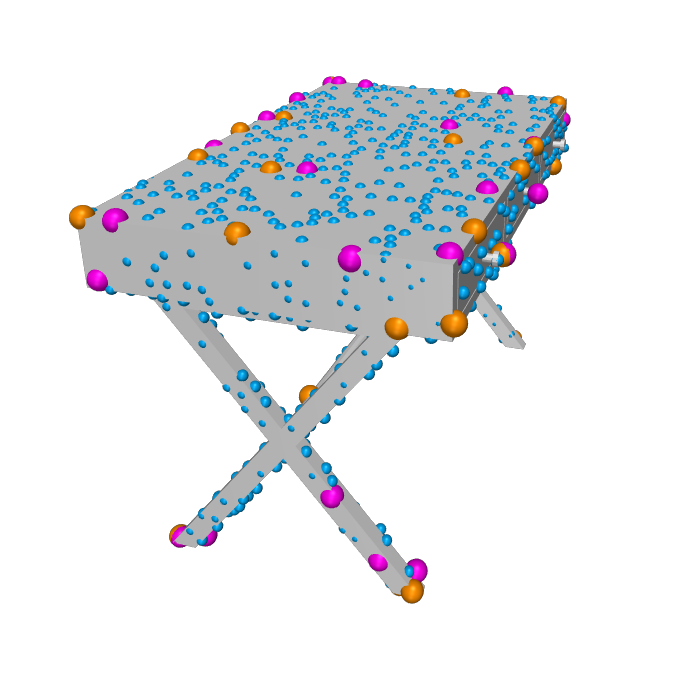} \\
\includegraphics[width=0.28\linewidth]{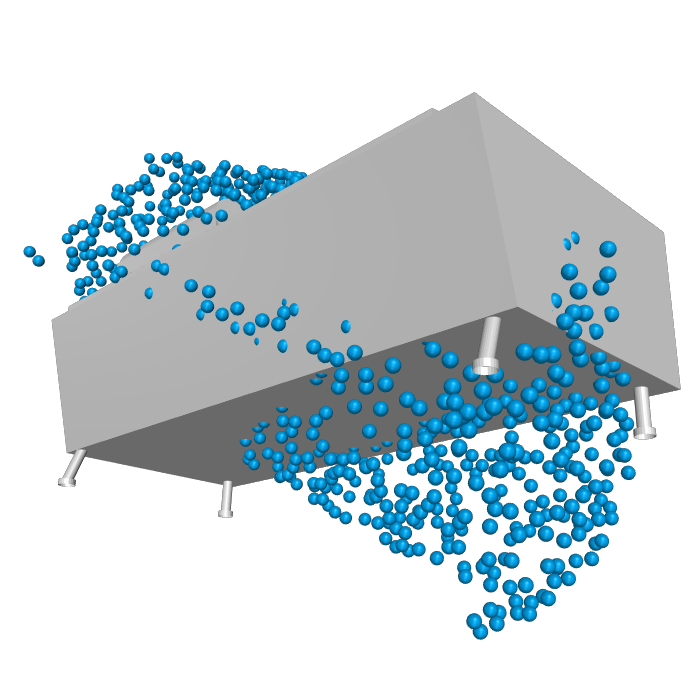} &
\includegraphics[width=0.28\linewidth]{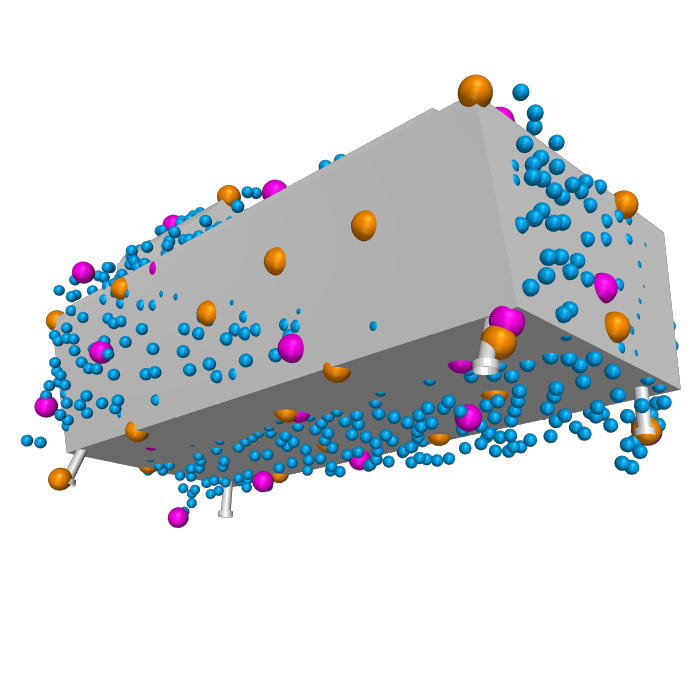} &
\includegraphics[width=0.28\linewidth]{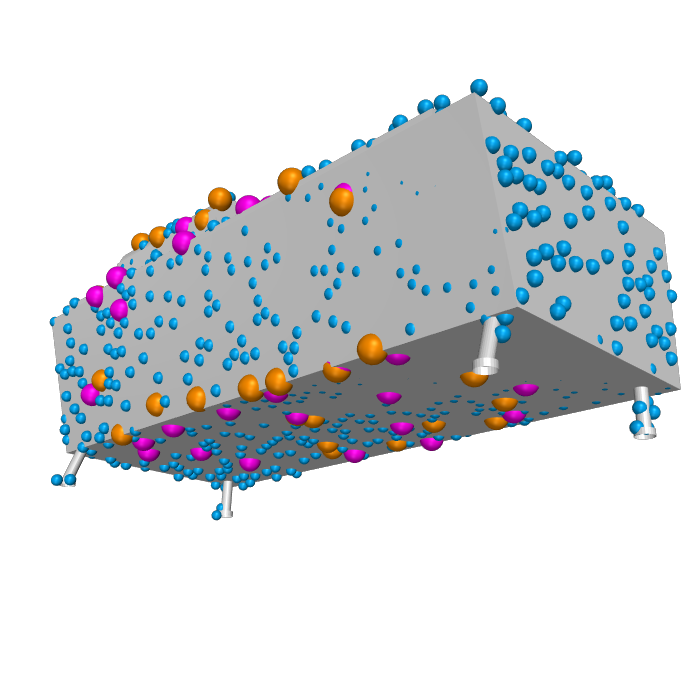} \\
\includegraphics[width=0.28\linewidth]{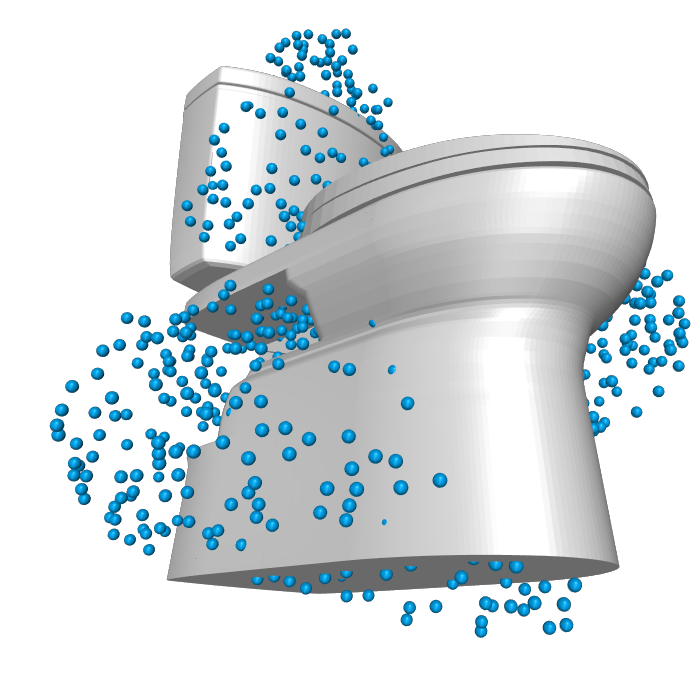} &
\includegraphics[width=0.28\linewidth]{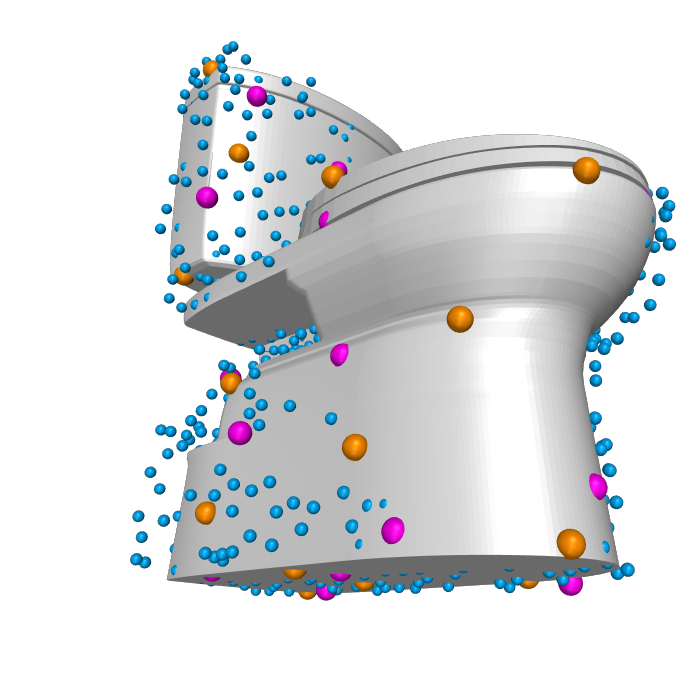} &
\includegraphics[width=0.28\linewidth]{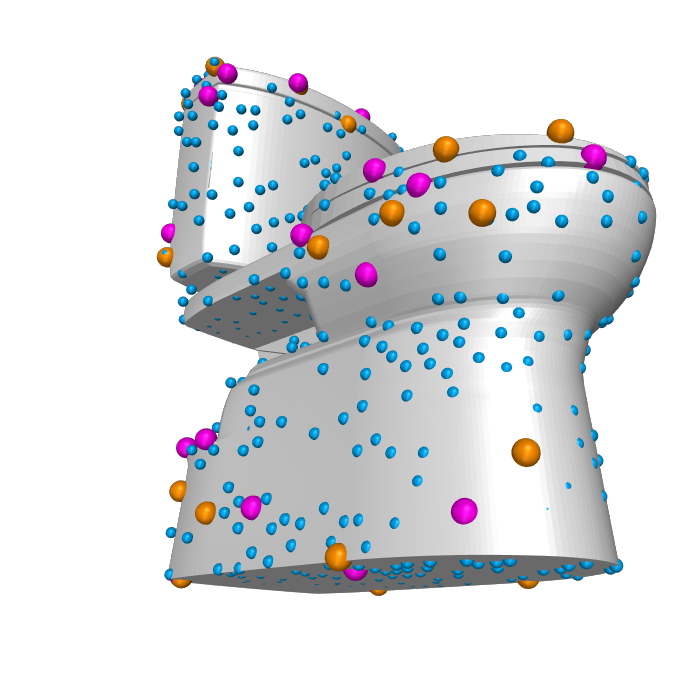} \\
\end{tabular}
\caption{{\bfseries Registration with sampled points for different shape categories.} Left column: unregistered source with 1024 points in Blue overlaid on the mesh model. Middle column: FPS registered results. Right column: \SPNet registered results. Sampled sets of 32 points from the template and source are illustrated in Orange and Magenta, respectively. Registration with \SPNet points yields better results than FPS.}
\label{fig:reg_additional}
\end{center}
\end{figure}

Registration is applied to different shape categories from ModelNet40. We present the results for Table, Sofa, and Toilet categories in Table~\ref{tbl:reg_additional_results}, and visualizations in Figure~\ref{fig:reg_additional}. Additional shape classes that we evaluated include Chair, Laptop, Airplane and Guitar. \SPNet achieves the best results compared to FPS and random sampling for all these categories.

\begin{table*}[htb!]
\begin{center}
\begin{tabular}{ l | c c c c c c c}
\hline
Sampling ratio & 2 & 4 & 8 & 16 & 32 & 64 & 128 \\
\hline
\hline
Random sampling & 1.03 & 2.59 & 5.29 & 9.99 & 18.53 & 34.71 & 63.09 \\
FPS & \textbf{0.46} & \textbf{1.5} & 3.3 & 6.42 & 11.78 & 22.23 & 43.49 \\
\SPNetwo & 0.53 & 1.64 & \textbf{3.14} & \textbf{4.83} & \textbf{6.85} & \textbf{7.2} & \textbf{9.6} \\
\hline
\end{tabular}
\end{center}
\caption{{\bfseries 
Sampling consistency between rotated point clouds.} The consistency is measured for the test split of Car category from ModelNet40. The results are multiplied by a factor of $10^3$. Lower is better. When the sampling ratio is small and many points are taken, \SPNet performs on par with the other methods. However, as it increases, \SPNet selects much more similar points than random sampling and FPS.}
\label{tbl:consistency}
\end{table*}

\begin{table*}[bth!]
\begin{center}
\begin{tabular}{ l | c c c | c c c | c c c}
\hline
Category & \multicolumn{3}{c}{\textit{Table}} & \multicolumn{3}{c}{\textit{Sofa}} & \multicolumn{3}{c}{\textit{Toilet}}\\
\hline
Sampling ratio & 8 & 16 & 32 & 8 & 16 & 32 & 8 & 16 & 32\\
\hline
\hline
Random sampling & 13.09 & 18.99 & 29.76 &
16.58 & 24.57 & 34.19 &
12.17 & 20.51 & 35.92 \\
FPS & 7.74 & 8.79 & 11.15 &
9.41 & 12.13 & 17.52 &
7.74 & 8.49 & 11.69 \\
\SPNet & \textbf{6.44} & \textbf{7.24} & \textbf{8.35} &
\textbf{8.56} & \textbf{10.8} & \textbf{10.97} &
\textbf{6.05} & \textbf{7.09} & \textbf{8.07} \\
\hline
\end{tabular}
\end{center}
\caption{{\bfseries Mean rotation error (MRE) with \SPNet for different shape categories.} MRE is reported in degrees. Lower is better. PCRNet is trained on complete point clouds of 1024 points from the Table, Sofa and Toilet categories of ModelNet40. The MRE is measured on the test split for different sampling methods. Utilizing \SPNet yields better results. With complete input, PCRNet achieves $6.08\degree$ MRE for Table, $7.15\degree$ MRE for Sofa, and $5.43\degree$ MRE for Toilet.}
\label{tbl:reg_additional_results}
\end{table*}

\section{Ablation study}  \label{sec:supp_ablation_study}

\subsection{Neighborhood size}
The neighborhood size $k = |\EuScript{N}_P(\mathbf{q})|$ is the number of neighbors in $P$ of a point $\mathbf{q} \in Q$, on which $\mathbf{q}$ is softly projected. This parameter controls the local context in which $\mathbf{q}$ searches for an optimal point to sample.

\begin{figure}[htb!]
\includegraphics[width=\columnwidth]{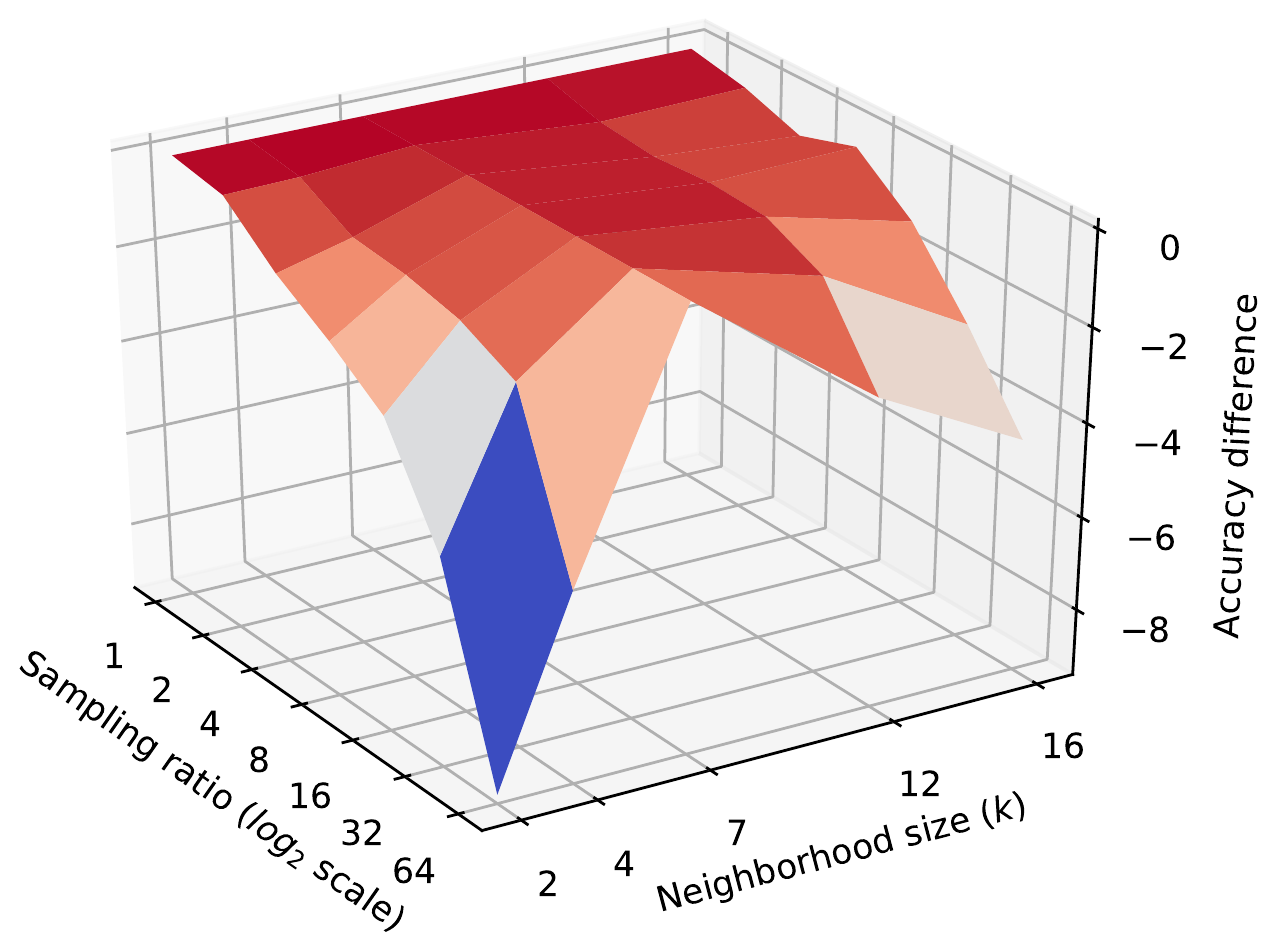}
\caption{{\bfseries The influence of different neighborhood sizes.} \SPNetwo-Progressive is trained for classification with different sizes $k$ for the projection neighborhood and evaluated on the test split of ModelNet40. We measure the accuracy difference for each sampling ratio with respect to the baseline of $k = 7$. Larger or smaller values of $k$ result in negative accuracy difference, which indicates lower accuracy.}
\label{fig:neighborhood size}
\end{figure}

We assess the influence of this parameter by training several progressive samplers for classification with varying values of $k$. Figure~\ref{fig:neighborhood size} presents the classification accuracy difference between \SPNetwo-Progressive trained with $k = 7$ and with $k \in \{2, 4, 12, 16\}$. The case of $k = 7$ serves as a baseline, and its accuracy difference is set to $0$. As shown in the figure, training with smaller or larger neighborhood sizes than the baseline decreases the accuracy. We conclude that $k = 7$ is a sweet spot in terms of local exploration region size for our learned sampling scheme.

\subsection{Additional loss terms}
As noted in the paper in section~\ref{sec:classification}, the average soft projection weights, evaluated on the test set of ModelNet40, are different than a delta function (see Figure~\ref{fig:weight_evolution}). In this experiment, we examine two loss terms, cross-entropy and entropy loss, that encourage the weight distribution to converge to a delta function.

For a point $\mathbf{q} \in Q$, we compute the cross-entropy between a Kronecker delta function, representing the nearest neighbor of $\mathbf{q}$ in $P$, and the projection weights of $\mathbf{q}$, namely, $\{w_i\}$, $i \in \EuScript{N}_P(\mathbf{q})$. The cross-entropy term takes the form:
\begin{equation} \label{eq:weight_cross_entropy}
H^c_P(\mathbf{q}) = -\sum_{i \in \EuScript{N}_P(\mathbf{q})}{\mathds{1}_{i^*}(i) log(w_i)} = -log(w_{i^*}),
\end{equation}

\noindent where $\mathds{1}_{i^*}(i)$ is an indicator function that equals $1$ if $i = i^*$ and $0$ otherwise; $i^* \in \EuScript{N}_P(\mathbf{q})$ is the index of nearest neighbor of $\mathbf{q}$ in $P$. The cross-entropy loss is the average over all the points in $Q$:
\begin{equation} \label{eq:loss_c}
\Lagr_{c}(Q,P) = \frac{1}{|Q|}\sum_{\mathbf{q} \in Q}{H^c_P(\mathbf{q})}.
\end{equation}

\noindent Similarly, the entropy of the projection weights for a point $\mathbf{q} \in Q$ is given by:
\begin{equation} \label{eq:weight_entropy}
H_P(\mathbf{q}) = -\sum_{i \in \EuScript{N}_P(\mathbf{q})}{w_i log(w_i)},
\end{equation}

\noindent and the entropy loss is defined as:
\begin{equation} \label{eq:loss_h}
\Lagr_{h}(Q,P) = \frac{1}{|Q|}\sum_{\mathbf{q} \in Q}{H_P(\mathbf{q})}.
\end{equation}

The cross-entropy and entropy losses are minimized when one of the weights is close to $1$, and the others to $0$. We add either of these loss terms, multiplied by a factor $\eta$, to the training objection of \SPNet (Equation~\ref{eq:loss_total_sample}), and train it for the classification task.

\begin{figure}[htb!]
\includegraphics[width=\columnwidth]{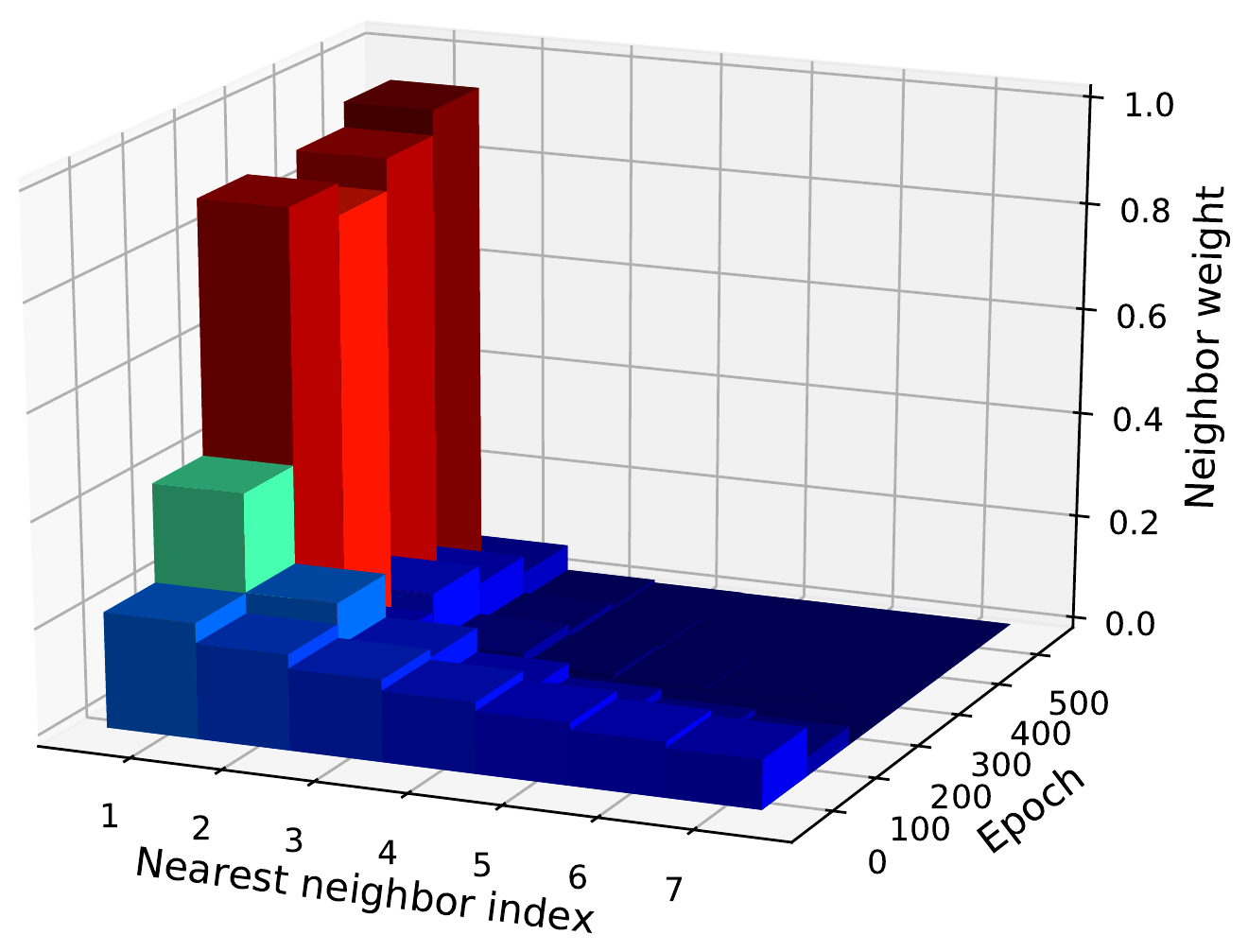}
\caption{{\bfseries Weight evolution with cross-entropy loss.} \SPNet is trained to sample 64 points for classification. A cross-entropy loss on the projection weights is added to its objective function. The weights are averaged on sampled point clouds from the test set of ModelNet40 after the first and every 100 training epochs. In these settings, most of the weight is given to the first nearest neighbor quite early in the training process.}
\label{fig:weight_evolution_supp}
\end{figure}

Figure~\ref{fig:weight_evolution_supp} presents the weight evolution for \SPNet that samples 64 points. It was trained with the additional cross-entropy loss, with $\eta = 0.1$. In these settings, the weights do converge quite quickly to approximately delta function, with an average weight of 0.94 for the first nearest neighbor at the last epoch. However, as Table~\ref{tbl:ablation_cross_entropy_loss} shows, this behavior does not improve the task performance, but rather the opposite.

The cross-entropy loss compromises the quest of \SPNet for optimal points for the task. Instead of exploring their local neighborhood, the softly projected points are locked on their nearest neighbor in the input point cloud early in the training process. We observed similar behavior when using the entropy loss instead of the cross-entropy loss. We conclude that the exact convergence to the nearest neighbor is not required. Instead, the projection loss (Equation~\ref{eq:loss_project}) is sufficient for \SPNet to achieve its goal - learning to sample an optimal point set for the task at hand.




\begin{table*}[htb!]
\begin{center}
\begin{tabular}{ l | c c c c c c c c}
\hline
Sampling ratio & 2 & 4 & 8 & 16 & 32 & 64 & 128 \\
\hline
\hline
\SPNet trained with cross entropy loss & 88.2 & 83.4 & 79.7 & 79.0 & 74.4 & \bfseries 55.5 & \bfseries 28.7 \\
\SPNet trained without cross entropy loss & \bfseries 88.4 & \bfseries 85.9 & \bfseries 83.8 & \bfseries 82.2 & \bfseries 80.1 & 54.0 & 23.2 \\
\hline
\end{tabular}
\end{center}
\caption{{\bfseries Ablation test for cross-entropy loss.} \SPNet is trained for classification, either with or without cross-entropy loss (Equation~\ref{eq:loss_c}). For each case, we report the classification accuracy on the test split of ModelNet40. Employing cross-entropy loss during training results in inferior performance for most of the sampling ratios.}
\label{tbl:ablation_cross_entropy_loss}
\end{table*}


\section{Mathematical aspects of soft projection} \label{sec:supp_mathematical_aspects}

\subsection{Idempotence}

Idempotence is a property of an operation whereby it can be applied several times without changing the obtained initial result. A mathematical projection is an idempotent operation. In the limit of $t \to 0$,  the soft projection becomes an idempotent operation. That is:
\begin{equation} \label{eq:idempotent}
\lim_{t \to 0} \sum_{i \in \EuScript{N}_P(\mathbf{q})}{w_i(t) \mathbf{p}_i} = \argmin_{\{\mathbf{p}_i\}} ||\mathbf{q}-\mathbf{p}_i||_2 = \mathbf{r}^*,
\end{equation}

\noindent which results in the definition of sampling in Equation~\ref{eq:hard_proj}. The proof of idempotence for the sampling operation is straightforward:
\begin{equation} \label{eq:idempotent_proof}
\argmin_{\{\mathbf{p}_i\}}
||\mathbf{r}^*-\mathbf{p}_i||_2 = \mathbf{r}^*.
\end{equation}

\subsection{Projection under the Bregman divergence}




The distance we choose to minimize between a query point $\mathbf{q} \in Q$ and the initial point cloud $P$ is the Squared Euclidean Distance (SED). However, SED is not a metric; it does not satisfy the triangle inequality. Nevertheless, it can be viewed as a Bregman divergence~\cite{chen2008metrics}, a measure of distance defined in terms of a convex generator function $F$.

Let $F: X \to \mathbb{R}$ be a continuously-differentiable and convex function, defined on a closed convex set $X$. The Bregman divergence is defined to be:
\begin{equation} \label{bregman}
D_{F}(\mathbf{p},\mathbf{q})=F(\mathbf{p})-F(\mathbf{q})-\langle \nabla F(\mathbf{q}),\mathbf{p}-\mathbf{q}\rangle.
\end{equation}

\noindent Choosing $F(\mathbf{x}) : \mathbb{R}^k \to \mathbb{R} = \left\lVert \mathbf{x}\right\rVert^2$, the Bregman divergence takes the form:
\begin{equation} \label{bregman_norm}
D_{F}(\mathbf{p}, \mathbf{q})=\left\lVert \mathbf{p}-\mathbf{q} \right\rVert^2.
\end{equation}

The projection under the Bregman divergence is defined as follows. Let \(\zeta \subseteq \mathbb{R}^k\) be a closed, convex set. Assume that ${F : \zeta \to \mathbb{R}}$ is a strictly convex function. The projection of $\mathbf{q}$ onto $\zeta$ under the Bregman divergence is:
\begin{equation} \label{bregman_proj}
\Pi_{\zeta}^{F}(\mathbf{q}) \triangleq \argmin_{\mathbf{r}\in \zeta} D_F(\mathbf{r}, \mathbf{q}).
\end{equation}

In our settings, the softly projected points are a subset of the convex hull of $\{\mathbf{p}_i\}$, $i \in \EuScript{N}_P(\mathbf{q})$. The convex hull is a closed and convex set denoted by $\zeta_{\mathbf{q}}$:


\begin{equation} \label{probsimplex}
\begin{split}
\zeta_{\mathbf{q}} = \left\{\mathbf{r}:\mathbf{r}=\sum_{i \in \EuScript{N}_P(\mathbf{q})}{w_i \mathbf{p}_i}, w_i \in [0,1], \sum_{i \in \EuScript{N}_P(\mathbf{q})}{w_i} = 1 \right\}
\end{split}
\end{equation}

In general, not all the points in $\zeta_\mathbf{q}$ can be obtained, because of the restriction imposed by the definition of $\{w_i\}$ in Equation \ref{eq:w_i}. However, as we approach the limit of ${t \to 0}$, the set $\zeta_\mathbf{q}$ collapses to $\{\mathbf{p}_i\}$. Thus, we obtain the sampling operation:
\begin{equation} \label{bregman_proj_relaxed}
\Pi_{\EuScript{N}_P(\mathbf{q})}^{F}(\mathbf{q}) \triangleq  \argmin_{\{\mathbf{p}_i\}}
D_F(\mathbf{p}_i, \mathbf{q}) = \mathbf{r}^*,
\end{equation}
\noindent as defined in Equation \ref{eq:hard_proj}.

\section{Experimental settings} \label{sec:supp_experimental_settings}

\subsection{Task networks}
We adopt the published architecture of the task networks, namely, PointNet for classification~\cite{qi2017pointnet}, PCRNet for registration~\cite{sarode2019pcrnet}, and point cloud autoencoder (PCAE) for reconstruction~\cite{achlioptas2018learning}. PointNet and PCAE are trained with the settings reported by the authors. Sarode \etal~\cite{sarode2019pcrnet} trained PCRNet with Chamfer loss between the template and registered point cloud. We also added a loss term between the estimated transformation and the ground truth one. We found out that this additional loss term improved the results of PCRNet, and in turn, the registration performance with sampled point clouds of \SPNetwo. Section~\ref{sec:supp_losses_metric_reg} describes both loss terms.

\subsection{\SPNet architecture}
\SPNet includes per-point convolution layers, followed by symmetric global pooling operation and several fully connected layers. Its architecture for different applications is detailed in Table~\ref{tbl:architectures}. For \SPNetwo-Progressive, the architecture is the same as the one in the table, with $m = 1024$ for classification and $m = 2048$ for reconstruction.


Each convolution layer includes batch normalization and ReLU non-linearity. For classification and registration, each fully connected layer, except the last one, includes batch normalization and ReLU operations. ReLU is also applied to the first two fully connected layers for the reconstruction task, without batch normalization.


\begin{table}[tb!]
\begin{center}
\begin{tabular}{ l c }
\hline
Task       &   \SPNet architecture   \\
\hline
\hline
&  $MLP(64, 64, 64, 128, 128)$    \\
Classification & max pooling      \\
& $FC(256, 256, 256, m \times 3)$ \\
\hline
&   $MLP(64, 64, 64, 128, 128)$   \\
Registration & max pooling        \\
& $FC(256, 256, 256, m \times 3)$ \\
\hline
& $MLP(64, 128, 128, 256, 128)$   \\
Reconstruction & max pooling      \\
& $FC(256, 256, m \times 3)$      \\
\hline
\end{tabular}
\end{center}
\caption{{\bfseries \SPNet architecture for different tasks.} $MLP$ stands for multi-layer perceptrons. $FC$ stands for fully connected layers. The values in $MLP(\cdot)$ are the number of filters of the per-point convolution layers. The values in $FC(\cdot)$ are the number of neurons of the fully connected layers. The parameter $m$ in the last fully connected layer is the sample size.}
\label{tbl:architectures}
\end{table}

\subsection{\SPNet optimization}
Table~\ref{tbl:hyper_parameters} presents the hyperparameters for the optimization of \SPNetwo. In progressive sampling for the classification task, we set $\gamma = 0.5$ and $\delta = 1/30$. The other parameter values are the same as those appear in the table. We use Adam optimizer with a momentum of $0.9$. For classification, the learning rate decays by a factor of $0.7$ every $60$ epochs. \SPNetwo-Progressive is trained with control sizes $C_s=\{2^l\}_{l=1}^{10}$ for classification and $C_s=\{2^l\}_{l=4}^{12}$ for reconstruction.


The temperature coefficient ($t$ in Equation~\ref{eq:w_i}) is initialized to $1$ and learned during training. In order to avoid numerical instability, it is clipped by a minimum value of $0.1$ for registration and $0.01$ for reconstruction.

We train our sampling method with a Titan Xp GPU. Training \SPNet for classification takes between $1.5$ to $7$ hours, depending on the sample size. The training time of progressive sampling for this task is about $11$ hours. The training time of \SPNet for registration takes between $1$ to $2.5$ hours. For the sample sizes of the reconstruction task, \SPNet requires between $4$ to $30$ hours of training, and \SPNetwo-Progressive requires about $2.5$ days.

\begin{table}[tb!] \label{hyper_parameters}
\begin{center}
\begin{tabular}{ l c c c c }
\hline
\ignorethis{Task} &  Classification   &   Registration    &     Reconstruction    \\
\hline
\hline
$k$              &   $7$             &   $8$         &       $16$            \\
$\alpha$         &   $30$            &   $0.01$      &       $0.01$          \\
$\beta$          &   $1$             &   $1$         &       $1$             \\
$\gamma$         &   $1$             &   $1$         &       $0$             \\
$\delta$         &   $0$             &   $0$         &       $1/64$          \\
$\lambda$        &   $1$             &   $0.01$      &       $0.0001$        \\
BS               &   $32$            &   $32$        &       $50$            \\
LR               &   $0.01$          &   $0.001$     &       $0.0005$        \\
TEs              &   $500$           &   $400$       &       $400$           \\
\hline
\end{tabular}
\end{center}
\caption{{\bfseries Hyperparameters.} The table details the values that we use for the training of our sampling method for different applications. BS, LR, and TEs stand for batch size, learning rate, and training epochs, respectively.}
\label{tbl:hyper_parameters}
\end{table}

\subsection{Losses and evaluation metric for registration} \label{sec:supp_losses_metric_reg}
Since the code of PCRNet~\cite{sarode2019pcrnet} was unavailable at the time of submission, we trained PCRNet with slightly different settings than those described in the paper, by using a mixture of supervised and unsupervised losses. 

The unsupervised loss is the Chamfer distance~\cite{achlioptas2018learning}:
\begin{equation} \label{eq:loss_reg1}
\begin{split}
\Lagr_{cd}(S,T) = \frac{1}{|S|}\sum_{\mathbf{s} \in S}{\min_{\mathbf{t} \in T}||\mathbf{s}-\mathbf{t}||_2^2} \\
+ \frac{1}{|T|}\sum_{\mathbf{t} \in T}{\min_{\mathbf{s} \in S}||\mathbf{t}-\mathbf{s}||_2^2},
\end{split}
\end{equation}

\noindent for a source point cloud $S$ and a template point cloud $T$. For the supervised loss, we take the quaternion output of PCRNet and convert it to a rotation matrix to obtain the predicted rotation $R_{pred}$. For a ground truth rotation $R_{gt}$,
the supervised loss is defined as follows:
\begin{equation} \label{eq:loss_reg2}
\Lagr_{rm}(R_{pred},R_{gt}) = ||R_{pred}^{-1}\cdot R_{gt} - I||_F^2,
\end{equation}

\noindent where $I$ is a $3 \times 3$ identity matrix, and $||\cdot||_F$ is the Frobenius norm. In total, the task loss for registration is given by $\Lagr_{cd}(S,T) + \Lagr_{rm}(R_{pred},R_{gt})$.

The rotation error $RE$ is calculated as follows~\cite{yuan2018pcn}:
\begin{equation} \label{eq:mre_metric}
RE = 2cos^{-1}(2 \langle q_{pred}, q_{gt} \rangle^2 - 1),
\end{equation}
\noindent where $q_{pred}$ and $q_{gt}$ are quaternions, representing the predicted and ground truth rotations, respectively. We convert the obtained value from radians to degrees, average over the test set, and report the mean rotation error.

\end{document}